\documentclass[10pt,journal,compsoc]{IEEEtran}
\usepackage[ruled,linesnumbered]{algorithm2e}
\usepackage{multirow}
\usepackage[table]{xcolor} 
\usepackage{amssymb}
\usepackage{amsfonts}
\usepackage{multicol}
\usepackage{graphicx}
\usepackage{booktabs}
\usepackage[T1]{fontenc}
\usepackage{float}
\usepackage[breaklinks=true,bookmarks=true,pagebackref=false,colorlinks,linkcolor=blue,anchorcolor=red,citecolor=blue,urlcolor=blue]{hyperref}
\usepackage[numbers,sort]{natbib}

\usepackage{comment}
\usepackage{subcaption}
\usepackage{amsmath}
\usepackage{algorithmic}
\usepackage{array}
\usepackage[switch]{lineno}
\usepackage{longtable}
\usepackage{xspace}
\usepackage{url}
\usepackage{overpic}
\usepackage{ragged2e}
\usepackage{framed}
\usepackage{enumitem}
\usepackage{balance}
\usepackage{makecell}
\usepackage{adjustbox}
\makeatletter
\DeclareRobustCommand\onedot{\futurelet\@let@token\@onedot}
\def\@onedot{\ifx\@let@token.\else.\null\fi\xspace}

\makeatother

\definecolor{darkgreen}{rgb}{0,0.7,0}
\definecolor{darkblue}{RGB}{31,119,180}
\definecolor{darkred}{RGB}{214,39,40}
\definecolor{mediumgray}{rgb}{0.5,0.5,0.5}
\definecolor{mediumteal}{rgb}{0,0.5,0.5}

\definecolor{ellisred}{rgb}{0.87,0.44,0.38} %
\definecolor{ellisgreen}{rgb}{0.69,0.90,0.52} %
\definecolor{elliscyan}{rgb}{0.29,0.77,0.74} %
\definecolor{ellisorange}{rgb}{0.89,0.55,0.28} %
\definecolor{ellisblue}{rgb}{0.41,0.61,0.86} %

\begin{document}
%
\title{Fully Unified Motion Planning for End-to-End Autonomous Driving
}

\author{Lin Liu, Caiyan Jia, Ziying Song, Hongyu Pan, Bencheng Liao,\\ Wenchao Sun, Yongchang Zhang,  Lei Yang, Yandan Luo
\IEEEcompsocitemizethanks{
\IEEEcompsocthanksitem  L. Liu,  C. Jia and Z. Song are with Beijing Key Laboratory of Traffic Data Mining and Embodied Intelligence, Beijing Jiaotong University. Email: $\{$songziying, 23120379, cyjia$\}$@bjtu.edu.cn.
\IEEEcompsocthanksitem H. Pan, B. Liao, W. Sun, Y. Zhang are with Horizon Robotics. Email:karry.pan@horizon.cc.

\IEEEcompsocthanksitem L. Yang is with School of Mechanical and Aerospace Engineering, Nanyang Technological University, Singapore. Email: lei.yang@ntu.edu.sg.

\IEEEcompsocthanksitem Y. Luo  is with School of Information Technology and Electrical Engineering, The  University of Queensland, Australia. Email: y.luo@uq.edu.au.

}

\thanks{Corresponding author: Caiyan Jia and Ziying Song.}
}

\markboth{Submitted to IEEE Transactions on Pattern Analysis and Machine Intelligence
}%
{Liu \MakeLowercase{\textit{et al.}}: 
Fully Unified Motion Planning for End-to-End Autonomous Driving
}
\IEEEtitleabstractindextext{%
\begin{abstract}
\justifying Current end-to-end autonomous driving methods typically learn only from expert planning data collected from a single ego vehicle, severely limiting the diversity of learnable driving policies and scenarios. However, a critical yet overlooked fact is that in any driving scenario, multiple high-quality trajectories from other vehicles coexist with a specific ego vehicle's trajectory. Existing methods fail to fully exploit this valuable resource, missing important opportunities to improve the models' performance (including long-tail scenarios) through learning from other experts. Intuitively, 
Jointly learning from both ego and other vehicles' expert data is beneficial for planning tasks. However, this joint learning faces two critical challenges. (1) Different scene observation perspectives across vehicles hinder inter-vehicle alignment of scene feature representations; (2) The absence of partial modality in other vehicles' data (e.g., vehicle states) compared to ego-vehicle data introduces learning bias. To address these challenges, we propose FUMP (Fully Unified Motion Planning), a novel two-stage trajectory generation framework. Building upon probabilistic decomposition, we model the planning task as a specialized subtask of motion prediction. Specifically, our approach decouples trajectory planning into two stages. In Stage 1, a shared decoder jointly generates initial trajectories for both tasks. In Stage 2, the model performs planning-specific refinement conditioned on an ego-vehicle's state. The transition between the two stages is bridged by a state predictor trained exclusively on ego-vehicle data. To address the cross-vehicle discrepancy in observational perspectives, we propose an Equivariant Context-Sharing Adapter (ECSA) before Stage 1 for improving cross-vehicle generalization of scene representations. Experimental results demonstrate that FUMP achieves state-of-the-art performance on planning tasks. Notably, FUMP achieves 36.1\% L2 error reduction (NuScenes), 28.6\% L2
error reduction (NuScenes-LT), +7.4\% DS (Bench2Drive), +3.7\% PDMS (NAVSIM), and 29.2\% lower collisions (Adv-NuScenes) compared to baselines. This effectively demonstrates the robustness and generalizability of the FUMP method in long-tail driving scenarios. This effectively demonstrates the robustness and generalizability of the FUMP method in driving scenarios, particularly in long-tail driving conditions. 
\end{abstract}

%
\begin{IEEEkeywords}
End-to-End Autonomous Driving, Motion Prediction, Plan, Bird's-Eye-View.

\end{IEEEkeywords}}

\maketitle

\IEEEdisplaynontitleabstractindextext

\IEEEpeerreviewmaketitle

\IEEEraisesectionheading{
\section{Introduction}\label{sec:introduction}
}
\IEEEPARstart{W}{i}th the rapid development of deep learning, autonomous driving technology is evolving from traditional handcrafted, modular architectures towards end-to-end architectures. Unlike conventional methods that process multiple sub-tasks separately, end-to-end autonomous driving systems strive to seamlessly integrate and jointly optimize multiple sub-tasks. These systems take sensor data as input and directly output future trajectories. Owing to their holistic optimization capabilities and the advantage of avoiding error accumulation across modules, end-to-end autonomous driving algorithms have achieved significant progress~\cite{uniad,sun2024sparsedrive,jiang2023vad,chen2024vadv2,cheng2024rethinking,guo2024uad,jia2023driveadapter,zheng2024genad,thiktwice}.
\begin{figure}[h]
    \centering
    \includegraphics[width=\linewidth]{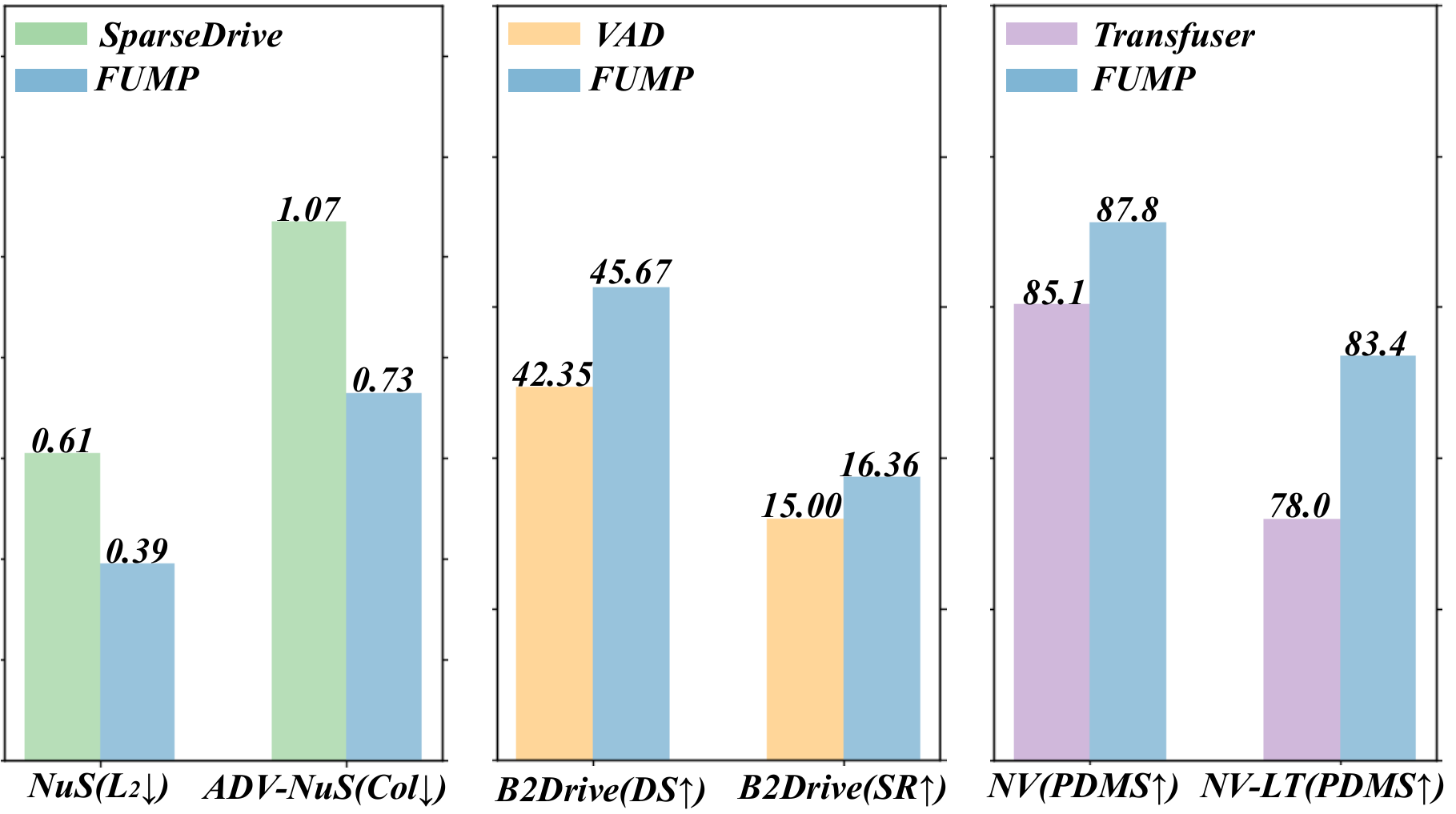}
    \caption{\textbf{The performance of FUMP}. FUMP significantly outperforms existing end-to-end baselines across NuScenes, Bench2Drive and NAVSIM, particularly on long-tail benchmark NAVSIM-LT (NV-LT). Notably, it achieves SOTA performance on the challenging collision dataset ADV-NuScenes (ADV-NuS)~\cite{challenger}. 'B2Drive' denotes Bench2Drive, 'NuS' represents NuScenes, and 'NV' stands for NAVSIM. }
    \label{fig:show}
 \end{figure}


While end-to-end autonomous driving systems~\cite{zheng2024genad,rhinehart2021contingencies, LookOut, CPOP, HGTP,su2024difsd,gameformer,k_game} demonstrate compelling performance in controlled environments, their real-world deployment faces two critical challenges, the prohibitive cost of collecting sufficient long-tail scenario data and The inherent limitation of single-expert trajectory supervision. These factors severely constrain the models' generalization capability in complex operational domain. The core limitation stems from a fundamental disparity in sample efficiency. Unlike perception tasks that can extract multiple learning samples from a single scene sample, current end-to-end paradigms only utilize one learnable expert trajectory (an ego vehicle's) per scene. This inefficiency necessitates exponentially growing data collection efforts to cover the required diversity of driving strategies and scenarios. A crucial observation reveals untapped potential: each scene sample inherently contains rich multi-agent motion data comprising trajectories from numerous "experts" (other vehicles). Exploiting these parallel expert demonstrations could simultaneously diversify the learnable strategy space and dramatically reduce data acquisition requirements.

Recent work explores three dominant paradigms for leveraging motion data (each with distinct tradeoffs), including (1) using other agents’ trajectories as interaction priors (limited to indirect planning benefits)~\cite{uniad,jiang2023vad,chen2024vadv2,sun2024sparsedrive,momad,diffusiondrive}, (2) jointly learning ego and non-ego trajectories (but requiring complex distillation or costly BEV transformations)~\cite{learningwatching,lav}, and (3) naively integrating motion prediction and planning modules (overlooking critical perspective discrepancies between vehicles)~\cite{zheng2024genad,su2024difsd,Int2Planner,HE-Drive}. In this work, we propose FUMP, a novel framework that enables the use of multi-agent trajectory data to train planning tasks. However, jointly learning from both motion trajectories (other vehicles) and planning trajectories (ego vehicles) presents two fundamental challenges in end-to-end autonomous driving.

\textbf{How to efficiently infer other vehicles' environmental observations from ego-centric perception?} Without access to other agents' raw sensor inputs, trajectory prediction must rely entirely on an ego vehicle's perception of the scene. Current methods predominantly model the relationships between environmental elements and an ego vehicle while neglecting inter-element dependencies. This limitation forces explicit coordinate transformations to construct agent-centric BEV representations for each neighboring vehicle, incurring substantial computational overhead. The root cause lies in existing scene representations' over-reliance on ego-centric reference frames, necessitating frequent coordinate conversions rather than directly capturing relative geometric relationships between traffic participants. To address this, we introduce an Equivariant Context-Sharing Scene Adapter (ECSA) as an initial module of FUMP to construct equivariant graph scene representation where each traffic element is modeled through its geometric relations to other elements, eliminating dependence on any fixed reference frame (e.g., ego coordinates). By explicitly encoding pairwise relative relationships, our method avoids the computational burden of perspective switching inherent in conventional approaches while preserving spatial consistency across all agents' viewpoints.

\textbf{How can a unified decoder jointly decode ego and other vehicles' trajectories given observational asymmetry in their states?} Planning tasks inherently incorporate an ego vehicle's state into decision-making, the inaccessibility of other agents' internal states creates fundamental challenges for unified learning of both ego and non-ego expert trajectories. To resolve this asymmetry, we propose a Unified Two-Stage Trajectory Decoder (UTTD) after the ECSA module, based on probabilistic decomposition. It reformulates the planning task as a specialized instance of motion prediction, and explicitly models trajectory generation as a two-stage process. In Joint Motion Policy Trajectory Generation stage, a state-agnostic shared decoder learns multi-expert trajectory distributions without requiring privileged vehicle state inputs. While in Conditional Plan Trajectory Refinement stage, an optimization submodule adapts trajectories by incorporating an ego vehicle's dynamics and intent through a self-state predictor trained exclusively on ego vehicle data. 
By this way, FUMP employs an ego-state predictor trained solely on ego-vehicle data, enabling seamless two-stage conversion. During inference, the framework optionally generalizes this predictor to other agents, further refining motion prediction.
\begin{figure}[t]
\centering
 \includegraphics[width=0.97\linewidth]{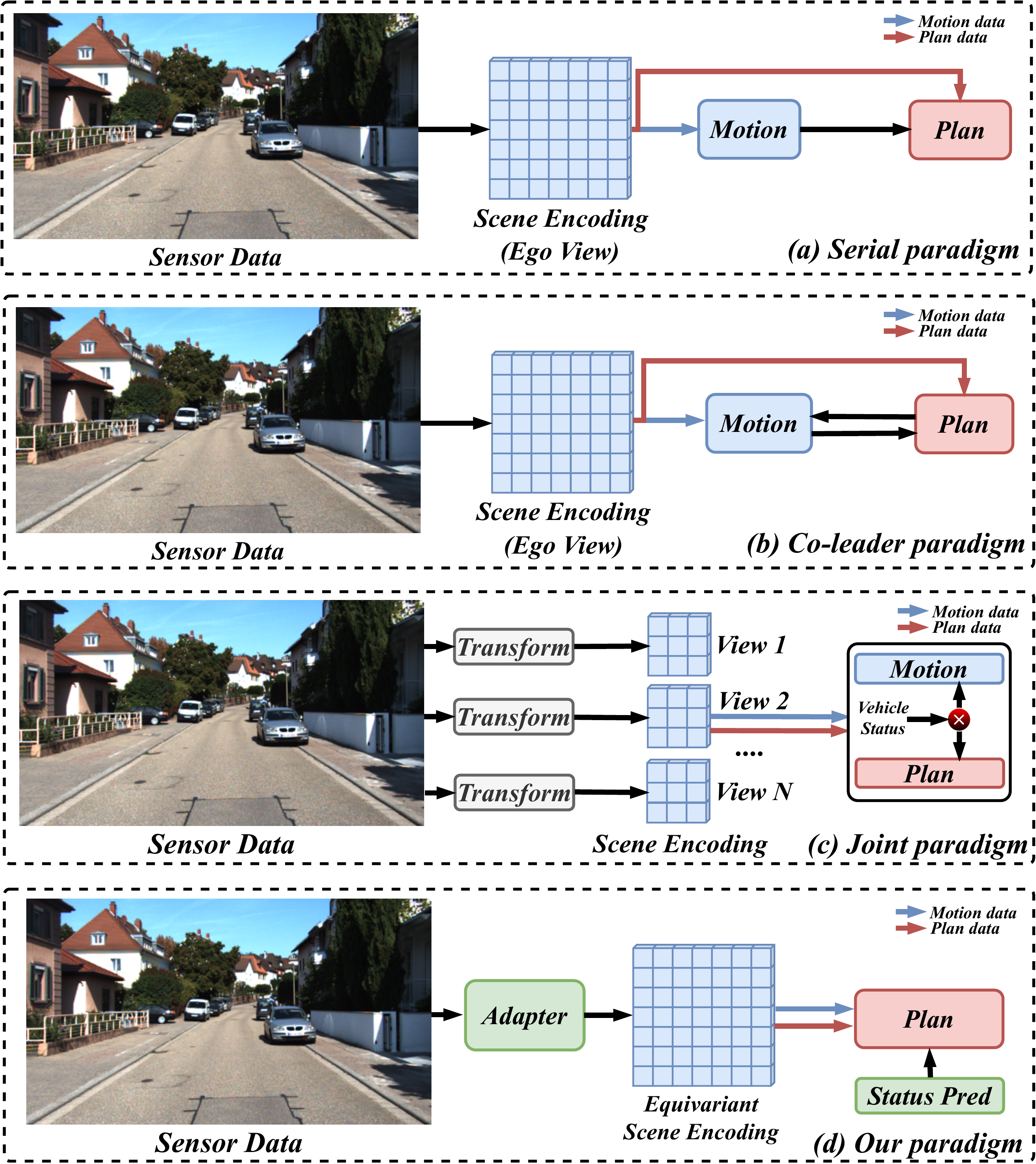}
\caption[ ]{\textbf{The paradigms for
leveraging motion data with planing tasks}. (a) \textbf{Serial Paradigm}, which ~\cite{Scept,grip,Densetnt,grip++,MultiMotion,Urtasun2020,uniad, chen2024vadv2, jiang2023vad} treats motion prediction outputs as prerequisites for planning tasks. The planning module benefits from motion data's future scenario modeling, thereby ensuring planning safety. (b) \textbf{Co-Leading Paradigm}, which~\cite{rhinehart2021contingencies, LookOut, CPOP, HGTP,su2024difsd, driveworld} independently constructs motion and planning modules while establishing iterative interaction mechanisms between them. Similar to (a), planning tasks inherently benefit from motion data's future scenario modeling. (c) \textbf{Joint Paradigm}, which~\cite{gameformer,zheng2024genad,lav,learningwatching} simultaneously outputs motion and planning results while attempting to learn both motion and planning data jointly. However, perspective differences in scene representation across vehicles and partial observability of other agents' states hinder unified learning. (d) \textbf{Our unified paradigm}, which seamless integrate the learning process of planning and motion data by introducing a perspective adapter. This adapter bridges viewpoint differences across vehicles and resolves the inaccessibility of other agents' states, facilitating joint learning of planning and motion.
}
\label{fig:motivation}
\end{figure}
Overall, our contributions are as follows.
\begin{itemize}
    \item  \textbf{FUMP} introduces a novel approach unifying motion and planning tasks, leveraging motion data to enhance planning performance and generalization.

    \item To bridge the gap between motion data and planning data, FUMP introduces two key components: (1) \textbf{Equivariant Context-Sharing Scene Adapter (ECSA)} that employs an Equivariant Graph Neural Network with Set Attention to extract geometrically consistent features from multi-agent interactions, eliminating perspective discrepancies; (2) \textbf{Unified Two-Stage Trajectory Decoder (UTTD)} that reformulates planning as a specialized motion sub-task through shared trajectory decoding and adaptive state estimation.
    
    \item Experiments on NuScenes~\cite{nuscenes}, ADV-NuScenes~\cite{challenger}, Bench2Drive~\cite{Bench2drive}, and NAVSIM~\cite{navsim} show FUMP’s effectiveness: 36\% L2 error reduction on NuScenes, 7.96\% Driving Score increase on Bench2Drive, 3.7\% PMDS improvement over TransFuser on NAVSIM, and 29.2\% collision rate reduction on ADV-NuScenes, achieving new SOTA, as shown in Fig.~\ref{fig:show}. The experimental results effectively demonstrate the robustness and generalizability of the FUMP method in driving scenarios, particularly on long-tailed datasets.
\end{itemize}

\begin{figure*}[!t]
    \centering
    \includegraphics[width=\linewidth]{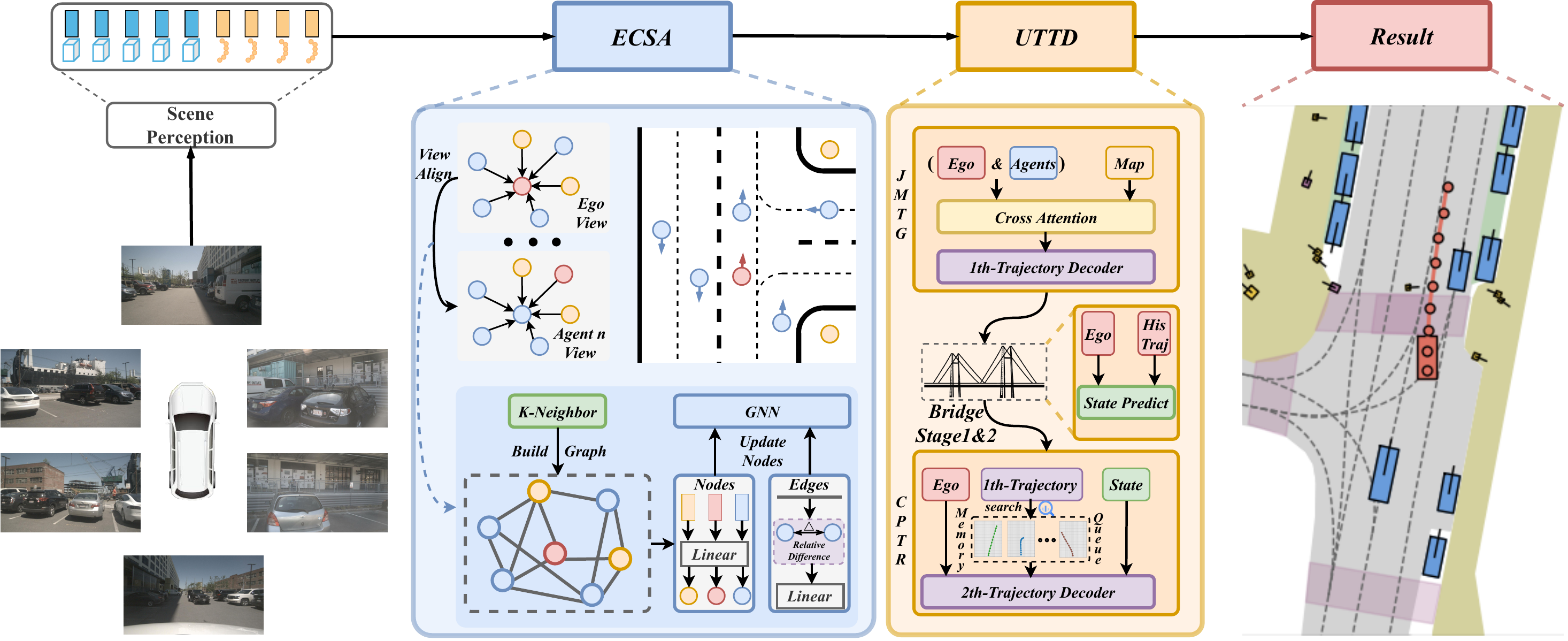}
    \caption{\textbf{The overall architecture of FUMP.} \textbf{(a)} Following SparseDrive, FUMP encodes multi-view images into feature maps, then represents a sparse scene as the composition of map instances and agent instances through sparse perception. \textbf{(b)} \textbf{ECSA} models traffic agents and map elements as graph nodes, dynamically grouping them into subgraphs while preserving global topology via inter-subgraph connections. An equivariant GNN then processes this hierarchical structure, learning multi-scale relationships through cross-group interactions to maintain geometric equivariance. Then, UTTD unifies motion and planning trajectory prediction through a two-stage framework: \textbf{(c) JMTG} and  \textbf{(d) CPTR}. During inference, an auxiliary vehicle state estimator is applied to other vehicles to address missing data modalities. Where \textbf{ECSA} denotes Equivariant Context-Sharing Scene Adapter, \textbf{JMTG} is Joint Motion-Policy Trajectory Generation and \textbf{CPTR} means Conditional Plan Trajectory Refinement.}
    \label{fig:framework}
\end{figure*}

\section{Related Work}\label{sec:related_work}
\subsection{End-to-end Autonomous Driving}
End-to-end autonomous driving directly maps raw sensor inputs data to future ego-vehicle trajectories, eliminating handcrafted rules. Early approaches~\cite{e_2_e_ICRA2018,e_2_e_Codevilla_2019_ICCV,e_2_e_TCP,e_2_e_Zhang_2021_ICCV_Coach,TransFuser} often employ process-implicit architectures without auxiliary task supervision (e.g., detection, online mapping, motion prediction), resulting in poor interpretability and optimization difficulties. Close-loop-based methods~\cite{TransFuser,jia2023driveadapter,thiktwice,gpt_driver,interfuser,shao2023reasonnet,xu2024m2da} learn vehicle actions in closed-loop simulations but struggle with the sim scene to real scene distribution gap, limiting their practical applicability.

Recent methods~\cite{chen2025ppad,doll2024dualad,guo2024uad,uniad,jiang2023vad,chen2024vadv2,Senna,gpt_driver,hedrive_zhangxingyu,fusionad,yang2024deepinteraction++,zhang2024graphad,zheng2024genad}, such as UniAD~\cite{uniad}, integrate perception, prediction, and planning modules for trajectory optimization, achieving state-of-the-art performance. VAD~\cite{jiang2023vad} introduces vectorized scene representations with explicit planning constraints, while VADv2~\cite{chen2024vadv2} enhances diversity via multi-modal trajectories. GraphAD~\cite{zhang2024graphad} models dynamic-static interactions via scene graphs, and PPAD~\cite{chen2025ppad} iteratively refines motion and planning outputs. SparseDrive~\cite{sun2024sparsedrive} incorporates symmetric sparse perception but lacks motion-planning interaction, a limitation addressed by DiFSD~\cite{su2024difsd}. Despite their advancements, these methods fail to leverage motion data for planning, missing opportunities to learn from out-of-distribution scenarios.

\subsection{Integration of Motion and Plannning}
While end-to-end autonomous driving systems typically treat planning as their central task, recent approaches increasingly recognize the influence of other agents' trajectory data (motion data) on planning performance. These methods attempt to leverage motion data to benefit the planner. In this section, we categorize and review these integration approaches into three paradigms based on their implementation as shown in Fig.~\ref{fig:motivation}:

1) Serial Integration. This paradigm~\cite{Scept,grip,Densetnt,grip++,MultiMotion,Urtasun2020,uniad, chen2024vadv2,jiang2023vad} processes motion prediction and planning as two distinct yet differentiable modules connected through an intermediate interface, enabling end-to-end training. In this architecture, the motion prediction module's outputs serve as informative priors for the planning module. This design allows the planner to leverage the motion predictor's scene forecasting capabilities, thereby enhancing the system's safety performance through more informed decision-making.

2) Co-leading Integration. 
The co-leading paradigms~\cite{rhinehart2021contingencies, LookOut, CPOP, HGTP,su2024difsd, driveworld} independently constructs motion and planning modules with iterative interaction, modeling bidirectional influences between SVs and EV planning. For instance, DTPP~\cite{dtpp} employs transformer-based conditional prediction for multi-agent coordination, while DIFSD~\cite{su2024difsd} iteratively refines planning constraints using motion outputs. Some approaches optimize EV (Ego Vehicle) decisions via single-frame multi-SV (Surrounding Vehicle) trajectories to minimize collision risks, whereas DriveWorld~\cite{driveworld} leverages world models to evaluate interactions. Although outperforming serial paradigms in motion-planning integration and future scene utilization, this framework still prevents planners from directly learning motion data.

3) Joint Integration. This paradigm focuses on joint learning from both motion and planning data. The LBW~\cite{learningwatching} method pioneers BEV-based planning but requires costly coordinate transforms for non-ego perspectives. LAV~\cite{lav} utilizes other agents' trajectories via augmented training signals, yet demands complex distillation for perspective consistency. Recent end-to-end approaches (GenAD~\cite{zheng2024genad}, Int2Planner~\cite{Int2Planner}, HE-Drive~\cite{HE-Drive}) employ unified decoders for joint SV/EV trajectory prediction. While their architectures enable joint motion-planning learning, this was not their primary design intention. More importantly, they fail to fully exploit motion data's potential for planning enhancement due to two fundamental oversights: (1) unresolved inter-vehicle perspective discrepancies, and (2) unobserved agent states. 

In contrast, this study aims to propose a scene-view adapter that efficiently aligns inter-agent perspectives with minimal computational cost. By reformulating planning as motion prediction sub-task, it targets to fully use unobservable agent states.

\section{Problem Formulation}
The motion prediction task aims to forecast the trajectory $\tau$ at time $t+1$ based on the accumulated perceptual observations $O_{m}$ up to time $t$. Crucially, as trajectory generation of a target vehicle is inherently coupled with vehicle state dynamics, this task implicitly depends on accurate estimation of the target vehicle's current state $s_{t}$ at time $t$. A straightforward example is that when estimating zero future trajectories for vehicles on both sides of the road, their current velocities are also implicitly estimated to be zero. The problem can thus be formulated as a joint conditional probability $P(\tau,s_{t}\mid O_{m})$, where trajectory prediction and state estimation are intrinsically coupled: the latent vehicle state $s_{t}$ governs motion dynamics, while a perceptual observation $O_{m}$ jointly constrains both the state and its future evolution.

The plan prediction task aims to predict the future trajectory $\tau$ at time $t+1$ based on the ego vehicle's current state $s_{t}$ and the accumulated perceptual observations $O_{p}$ up to time $t$, which can be probabilistically modeled as $P(\tau \mid O_{m}, s_{t})$. Under the assumption that the perceptual observations from both ego and other vehicles exhibit no significant perspective discrepancy $O_{m} \approx O_{p} \approx O$, we can establish a theoretical connection between motion prediction and plan prediction tasks. 
\begin{equation}
  \begin{aligned}
     P(\tau,s_{t}\mid O) = P(\tau \mid O, s_{t}) \cdot P(s_{t} \mid O).\\
  \end{aligned}
\end{equation}

Thus, the plan task can be viewed as a specialized probabilistic sampling of the motion task based on $P(s_{t} \mid O)$. This unification of motion and plan prediction can be decomposed into two key challenges.

1) Perception Alignment: Ensuring $O_{m} \approx O_{p} \approx O$ presents a fundamental challenge since other vehicles' sensor data is inherently unobservable. Only ego-vehicle perception data $O_{p}$ can be accessed, requiring learned perspective transformation of ego-centric observations to gain an other-vehicle view $O_{m}$. This necessitates the design of a view-unification module to achieve $O_{m} \approx O_{p} \approx O$ in the feature space.

2) Hierarchical Task Unification: The planning task can be derived through targeted sampling from motion predictions $P(\tau,s_{t}\mid O_{m})$ based on $P(s_{t} \mid O)$. Thus, both tasks share a common motion decoder, enabling efficient knowledge transfer. Therefore, the plan task's prediction of ego vehicles' state $P(s_{t} \mid O)$ can directly transform motion predictions into planning decision, creating a cohesive pipeline.
\begin{equation}
  \begin{aligned}
     P(\tau \mid O, s_{t}) = P(\tau,s_{t}\mid O) /  P(s_{t} \mid O).\\
  \end{aligned}
\end{equation}
We then formulate deep neural networks to model these probabilistic distributions as below.
\begin{equation}
  \begin{aligned}
f_{\theta}: \mathcal{O}  & \rightarrow \mathcal{P}(\mathcal{T},\mathcal{S}), \mathbf{O} \mapsto f_{\theta}\left(\tau \mid \mathbf{O}, \mathbf{s}_{t}\right), \\
g_{\phi}:  & \mathcal{O}  \rightarrow \mathcal{P}(\mathcal{S}),  \mathbf{O} \mapsto g_{\phi}\left(\mathbf{s}_{t} \mid \mathbf{O}\right), \\
h_{\psi}: \mathcal{O} \times \mathcal{S} & \rightarrow \mathcal{P}(\mathcal{T}),  \left(\mathbf{O}, \mathbf{s}_{t}\right) \mapsto h_{\psi}\left(\tau \mid \mathbf{O}, \mathbf{s}_{t}\right). \\
\end{aligned}
\end{equation}
Namely, we use the neural networks $f_{\theta}\left(\tau  \mid \mathbf{O}, \mathbf{s}_{t}\right)$, $g_{\phi}\left(\mathbf{s}_{t} \mid \mathbf{O}\right)$ and $h_{\psi}\left(\tau \mid \mathbf{O}, \mathbf{s}_{t}\right)$ to approximate the probability distributions $P\left(\tau, \mathbf{s}_{t} \mid \mathbf{O}\right)$, $P\left(\mathbf{s}_{t} \mid \mathbf{O}\right)$, $P\left(\tau \mid \mathbf{O}, \mathbf{s}_{t}\right)$, respectively. Where $\mathcal{P}(\cdot)$ denotes a probability measure space, and $\theta$, $\phi$, $\psi$ are neural parameters linked through the following diffeomorphic constraint,
\begin{equation}
\begin{aligned}
h_{\psi}\left(\tau \mid \mathbf{O}, \mathbf{s}_{t}\right)=\frac{f_{\theta}\left(\tau, \mathbf{s}_{t} \mid \mathbf{O}\right)}{g_{\phi}\left(\mathbf{s}_{t} \mid \mathbf{O}\right)}.
\end{aligned}
\end{equation}



\section{Method}\label{sec:method}
Fig.~\ref{fig:framework} illustrates the architecture of the proposed FUMP (a fully unified motion planning framework), which introduces a novel end-to-end paradigm integrating motion prediction and trajectory planning. This framework enables sufficient knowledge transfer from motion data to enhance planning performance through joint learning. The core innovations consist of two principal components. 1) \textbf{Equivariant Context-Sharing Scene Adapter} (Sec~\ref{sec:ECSA}), which represents traffic agents and map elements as graph nodes, dynamically partitioning them into subgraph groups and preserving global topology through inter-subgraph connections. The equivariant graph neural network subsequently processes the hierarchical graph structure, where cross-group message passing ensures geometric equivariance in scene representations while maintaining cross-perspective generalization. 2) \textbf{Unified Two-Stage Trajectory Decoder} (Sec~\ref{sec:UTTD}), which further comprises two sub-modules, \textit{Joint Motion-Policy Trajectory Generation} (Sec~\ref{sec:LRS}) and \textit{Conditional Plan Trajectory Refinement} (Sec~\ref{sec:LRS}). Where the former produces multi-modal trajectory hypotheses conditioned on structured scene graphs, the latter executes context-aware trajectory optimization by integrating ego-state dynamics. To enable unified prediction, an auxiliary Ego-State Predictor (ESP) is trained using historical observations and environmental contexts. During inference, this probabilistic estimator ESP is symmetrically applied to vehicle agents, achieving unified motion forecasting and ego-planning within the same geometric learning framework.

\begin{figure}[h]
    \centering
    \includegraphics[width=\linewidth]{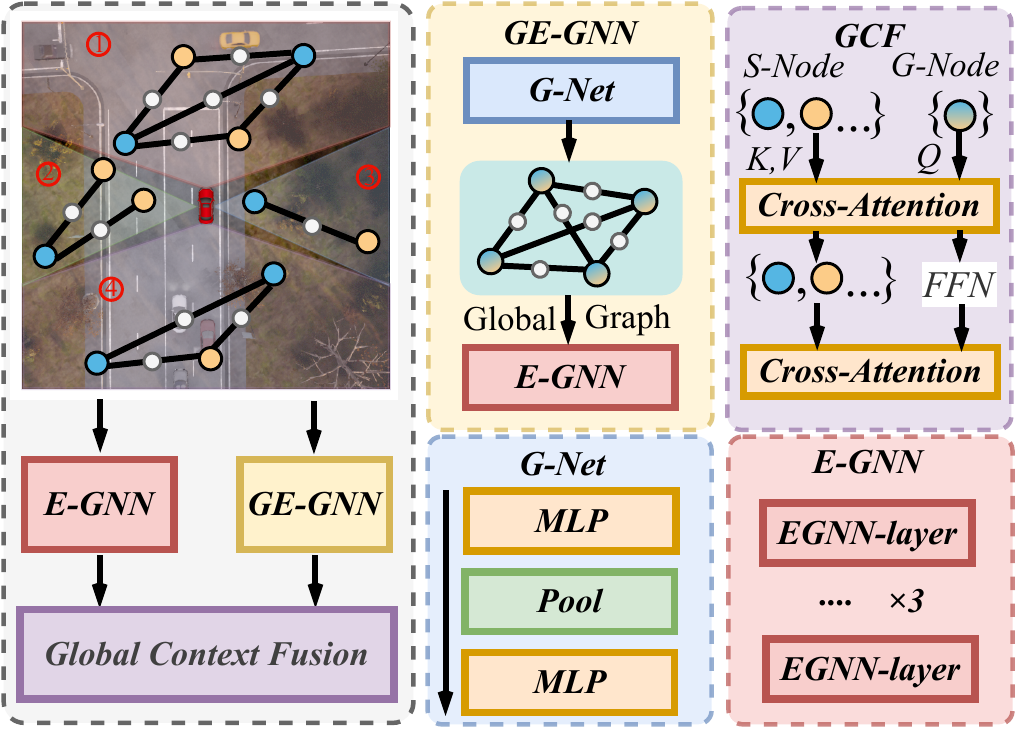}
     \caption{\textbf{The details of ECSA module}. Where the EGNN-layer follows EGNN from ~\cite{egnn}, GCF denotes Global Context Fusion module.}
    \label{fig:dataset_fs}
\end{figure}

\subsection{Equivariant Context-Sharing Scene Adapter}\label{sec:ECSA}
The core limitation stems from encoding scenes in fixed reference frames, requiring costly coordinate transformations when the reference changes. Inspired by EGNN's relational inductive bias (where node interactions depend solely on relative attributes), ECSA models traffic scenes as geometric graphs, which construct traffic elements (agents/map features) as nodes and constructs Invariant relative relationships as edges. Since these geometric relationships remain identical across any reference frames, ECSA achieves implicit perspective alignment without explicit transformations. To enhance the model’s processing efficiency, FUMP constructs hierarchical subgraphs to learn scene representations. The scene is represented as a global graph $G_{g}$ with $N_{zone}$ subgraphs $G_{s}$, where $N_{zone}$ is 4.

Every scene is partitioned into four regions based on the human driver’s attention range and the ego vehicle’s heading angle: Forward Zone ($+30^\circ$ to $+150^\circ$), Lateral Zones ($-30^\circ$ to $+30^\circ$ and $150^\circ$ to $210^\circ$), and the rearward zone ($-30^\circ$ to $-150^\circ$). FUMP constructs a subgraph for each zone, where edges are established by computing pairwise distances between graph nodes and connecting each node to its $K$-nearest neighbors. For each subgraph $G_{s} = (V_{s}, E_{s})$ , node features $v_{i} \in V_{s}$ are initialized from $F_{a}$ and $F_{m}$ , where $F_{a}$ and $F_{m}$ denote the encoded features of agents and map elements extracted by the perception module, respectively. And each edge features $e_{ij} \in E_{s}$ are defined according to the following formulation.
\begin{equation}
  \begin{aligned}
    r_{ij}^{s} &= \left [ \left \| \phi_{e}(c_{i}) -  \phi_{e}(c_{j})\right \|,vel_{i} - vel_{j}, cls_{i}^{v}, cls_{j}^{v}  \right ], \\
    e_{ij}^{s} &= MLP(r_{ij}).
\\  \end{aligned}
\end{equation}
Here, $c_{i}$ and $c_{j}$ denote the coordinates of traffic elements associated with nodes $i$ and $j$ respectively in the ego-vehicle coordinate frame. The positional encoding operation is represented as $\phi_{e}$. The node features of node $i$ and node $j$ are denoted by $v_{i}^{l}$ and $v_{j}^{l}$. $vel$ represents the node velocity, where the $vel$s of map nodes are set to zero and $cls_{i}^{v}$ denotes the node category. Critically, edges between nodes are constructed solely based on their relative attribute quantities that are both coordinate-invariant and viewpoint-agnostic. This guarantees consistent inter-element geometric relationships regardless of observation perspective. Based on EGCL~\cite{LaneGP}, the node features are updated as follows.
\begin{equation}
\begin{aligned}
    m_{ij} &= g_{e}(v_{i}^{l}, v_{j}^{l}, e_{ij} ), \\
    v_{i}^{l+1} &= v_{i}^{l} + C\sum_{j\ne i} c_{ij} \cdot g_{x}(m_{ij}),\\
    v_{i}^{l+1} &= g_{h}(v_{i}^{l}, \sum_{j} m_{ij}).
\end{aligned}
\label{egnn}
\end{equation}
Here, the edge operation $g_{e}$ and node operation $g_{h}$ are approximated by multilayer perceptrons (MLPs), while $g_{x}$ is also an MLP. ECSA implicitly achieves perspective alignment by modeling node-to-node relative relationships that are inherently invariant to reference frame transformations.

After updating the node features within each subgraph, ECSA proceeds to construct relationships between subgraphs. The global graph adopts full connectivity. For global graph $G_{g} = (V_{g}, E_{g})$ , For the global graph, its node features $v_{i} \in V_{g}$ are aggregated from all nodes in its subgraphs using PointNet, and its edge attributes $e_{ij} \in E_{g}$ are defined as follows: 
\begin{equation}
  \begin{aligned}
  V_{g} &= PointNet(V_{S}) \\
    r_{ij}^{g} &= \left [ \left \| \phi_{e}(c_{i}) -  \phi_{e}(c_{j})\right \|,vel_{i} - vel_{j} \right ], \\
    e_{ij}^{g} &= MLP(r_{ij}^{g}).
\\  \end{aligned}
\end{equation}
Here, 
$c$ and $vel$ of each global node are computed as the mean and vector sum of all corresponding nodes in their subgraphs, respectively. Nodes and edges updates in the global graph are performed according to the operations defined in Eq.~\ref{egnn}.

After gaining updated subgraph nodes $V$ and global graph nodes $GV$, FUMP focuses on propagating the learned global relationships back to local subgraphs. For each subgraph’s node set $V_{i}$ and the global node $GV_{i}$, FUMP utilizes a set self-attention mechanism to propagate the learned global relations back to the subgraph nodes, facilitating both the updating of global relationships and the intra-subgraph interactions among nodes, as shown in Eq.~\ref{set_attn}.
\begin{equation}
  \begin{aligned}
    \hat{GV_{i}} &= CA(GV_{i},V_{i},V_{i}) \in \mathbb{R}^{1 \times d}, \\
    &\hat{H_{i}} = FFN(\hat{GV_{i}}),\\
    \hat{V_{i}} &= CA(V_{i},\hat{H_{i}},\hat{H_{i}}) \in \mathbb{R}^{k \times d}.
\\  \end{aligned}
\label{set_attn}
\end{equation}
Where $k$ denotes the number of nodes within a subgraph, $d$ represents the feature dimensionality, and CA denotes the Cross-Attention module.

\subsection{Unified Two-Stage Trajectory Decoder}\label{sec:UTTD}
To enable the sharing of trajectory decoder parameters across motion and planning tasks with heterogeneous data modalities, thereby facilitating knowledge transfer, FUMP introduces a Unified Two-Stage Trajectory Decoder architecture.

\subsubsection{Joint Motion-Policy Trajectory Generation}\label{sec:PGS}
\textit{\textbf{In Stage I, }}FUMP refrains from incorporating an ego vehicle state, instead predicting all potential trajectories for an ego vehicle in the same manner as it forecasts those for surrounding vehicles. This process, termed \textbf{Joint Motion-Policy Trajectory Generation} stage, leverages cross-attention mechanisms between a motion query $Q_{m}\in \mathbb{R}^{N_{m} \times d}$ and a plan query $Q_{p}\in \mathbb{R}^{N_{p} \times d}$, as well as agent and map elements, to achieve comprehensive information fusion. The integrated $Q_{m}$ and $Q_{p}$ features are processed by the first-stage trajectory decoder to produce the preliminary trajectory estimation.
\begin{equation}
  \begin{aligned}
    \{Q_{m}, Q_{p}\} = CA(\{Q_{m}, Q_{p}\},V,V) \in \mathbb{R}^{ (N_{p} + N_{m}) \times K \times d},\\
    \{\tau_{m}^{\mathbb{P}}, \tau_{P}^{\mathbb{P}}\} = \textit{TDC}(\{Q_{m}, Q_{p}\}) \in \mathbb{R}^{ (N_{p} + N_{m}) \times t_{traj} \times K \times 2}.
    \end{aligned}
\label{cross_attn}
\end{equation}
Here, TDC denotes the Trajectory Decoder module. The trajectory length, denoted as $t_{traj}$, is configured as 6 in the decoding stage. $K$ denotes the num of trajectory proposals and is set to 6. In the first stage of trajectory supervision, $L_{2}$ loss is applied to motion-task trajectories and GT (the ground truth) $\tau_{m}^{gt}$, while planning-task trajectories adopt a distribution-based metric and $L_{2}$ loss. 

The GT $\tau_{m}^{gt}$ based on distributional supervision is derived as follows. At time $t$, with the ego vehicle’s center $p_{ego}$ as the origin, a semicircle of radius $vel \cdot t_{traj}$ is drawn along the motion direction. Its intersection with the center lane $\mathcal{L}_c$ defines potential future positions, used as planning GT $\tilde{\tau_{p}^{gt}}$. Formally,
\begin{equation}
  \begin{aligned}
  \tilde{\tau}_{p}^{gt} = \left\{ \mathbf{p} \, \bigg| \, \mathbf{p} \in \mathcal{L}_c \, , \|\mathbf{p - p_{ego}}\|_2 = vel \cdot t_{traj} \right\}.
  \end{aligned}
\label{fake_plan_gt}
\end{equation}
\begin{equation}
  \begin{aligned}
  loss_{1th}^{plan} = \mathbb{E}_{k<K}\left(\min _{k^{\prime}<K_{gt}}\left\|\hat{\tau}_{p, k}^{\mathbb{P}}-\tilde{\tau}_{p, k^{\prime}}^{gt}\right\|_{2}\right)<d.
  \end{aligned}
\label{fake_plan_loss}
\end{equation}
Where $K_{gt}$ denotes the number of ground truth $\tilde{\tau}_{p}^{gt}$, $\tilde{\tau}_{p,k}^{gt}$ represents the $k$-th predicted value in $\tilde{\tau}_{p}^{gt}$, and $k^{\prime}$ signifies the $k^{\prime}$-th ground truth in $\tilde{\tau}_{p}^{gt}$.
\subsubsection{Conditional Plan Trajectory Refinement}\label{sec:LRS}


\textit{\textbf{In Stage II, }}FUMP performs fine-grained trajectory prediction modeling. Stage I establishes unified motion-planning representations, exposing motion data to planning tasks. However, the model should prioritize the samples which are similar to long-tail planning samples from motion data rather than redundant simple motion samples. Inspired by world model~\cite{world_model}, FUMP incorporates a memory module that actively identifies and stores motion samples potentially beneficial for planning tasks, subsequently leveraging these memorized patterns to guide planning decisions. FUMP finds an intrinsic overlap between long-tail planning scenarios (e.g., lane changes, turns) and challenging motion prediction cases.  


This observation motivates our memory-based approach that prioritizes high-loss motion samples within each training batch. FUMP implements the memory module as a fixed-capacity array-based queue that stores the top $N$ highest-loss motion samples per training batch. Crucially, FUMP's queue employs a dynamic replacement strategy based on exponential moving averages, rather than FIFO updates. Motion samples are enqueued if their loss exceeds a dynamic threshold. The threshold $\epsilon$ is dynamically adjusted via EMA over time in the following.
\begin{equation}
  \begin{aligned}
   \epsilon_{t} = \gamma \cdot \epsilon_{t-1} + \left ( 1 - \gamma \right ) \cdot loss_{t}.
  \end{aligned}
\label{ema}
\end{equation}
Here, the scaling factor $\gamma$ is set to 0.2 and $loss_{t}$ denotes the mean loss over motion samples in the current batch at time $t$. When the loss of a single sample exceeds a threshold $\epsilon_{t}$, it is pushed into the queue. If the queue is full, the new sample replaces the one with the lowest loss. Additionally, during each update, a random sample in the queue is replaced to ensure persistent memorization of hard motion samples as the network converges. The queue length $Length$ is maintained at 700. 

To effectively leverage hard samples for guiding planning decisions, FUMP performs trajectory matching between the first-stage trajectory output and hard samples in the queue. The matched sample trajectory $\tau_{m}^{\ast}$ is then encoded into the planning query $Q_{p}$ as shown in Eq~\ref{motion_sample},
\begin{equation}
  \begin{aligned}
   Q_{p} = Fusion(Q_{p},\psi_{e}(\tau_{m}^{\ast})),
  \end{aligned}
\label{motion_sample}
\end{equation}
where $\psi_{e}$ and $Fusion$ are multi-layer perceptions (MLPs). This design is motivated by the observation that motion patterns of other agents may be reusable for an ego vehicle’s future planning.

Then, during the training, leveraging the multi-modal trajectory predictions and corresponding queries from the first phase, the ego-vehicle state serves as the pivotal element for prediction, enabling precise trajectory optimization through the integration of the following components.
\begin{equation}
  \begin{aligned}
   {\tau^{\prime}}_{p}^{\mathbb{P}} = \textit{TDC}&(\{Q_{p},m\odot ST, \tau_{p}^{\mathbb{P}}\}) \in \mathbb{R}^{N_{p} \times K \times t_{traj} \times 2},\\
   STP &= \min_f \left\| f(Q_{p}, \tau_{p}^{his}) - ST \right\|_{2}.
  \end{aligned}
\label{plan_2th}
\end{equation}
In this context, $ST$ denotes the ego vehicle’s state. The random mask $m \in \{0,1\}$ is applied during training with a 6.25\% probability of being zero. This design serves as a dropout-like regularization mechanism to prevent the model from over-relying on ego-vehicle states and developing shortcut learning behaviors~\cite{bevplanner}. Notably, during the training process, an additional STP (\textit{State Predictor}) is trained. This predictor takes the vehicle’s historical trajectory $\tau_{p}^{his}$ and features $Q_{p}$ as input, outputs the predicted vehicle state, and is supervised by the ground truth of state (denoted as $ST$ in the above equation) for optimization. During the testing phase, the output $ST_{p}$ of trained STP module $STP(Q_{p},\tau_{p}^{his})$ replaces ground-truth vehicle state $ST$ for trajectory prediction. Given the motion task's vehicle features and a historical trajectory input, our two-stage trajectory prediction framework operates as follows to obtain the trajectory of refined ego vehicle ${\tau^{\prime}}_{p}^{\mathbb{P}}$.
\begin{equation}
  \begin{aligned}
   {\tau^{\prime}}_{p}^{\mathbb{P}} = \textit{TDC}&(\{Q_{p},STP(Q_{p},\tau_{p}^{his}), \tau_{p}^{\mathbb{P}}\}) ,\\
   {\tau^{\prime}}_{p}^{\mathbb{P}}& \in \mathbb{R}^{N_{p} \times K \times t_{traj} \times 2}.\\
  \end{aligned}
\label{plan_2th}
\end{equation}


\section{Experiments}\label{sec:exps}
In this section, we  comprehensively validate our proposed framework FUMP both on the commonly used open-loop dataset NuScences, and the well-known close-loop datasets including Bench2Drive and NAVSIM.

\subsection{Experiments Setup}

\subsubsection{Datasets}
\noindent\textbf{Open-Loop.} First, FUMP conducts comprehensive experiments on the widely adopted open-loop benchmark dataset, NuScenes~\cite{nuscenes} for planning task. The NuScenes dataset comprises 1,000 driving sequences, each spanning 20 seconds and containing 40 annotated key frames. Each data sample includes six images captured by surrounding cameras, providing a 360° field of view. Point clouds are collected by LiDAR and radar sensors. In our experiments, we exclusively utilized image data as model inputs, omitting LiDAR data.

\noindent\textbf{Close-Loop.} FUMP is then evaluated on the closed-loop benchmark dataset, Bench2Drive~\cite{Bench2drive} and NAVSIM. Bench2Drive is a large-scale expert dataset comprising both open-loop and closed-loop evaluations, featuring three distinct data partitions: \textit{mini} (10 clips), \textit{base} (100 clips), and \textit{large} (10,000 clips). The open-loop training data consists of two million fully annotated frames, with each key sample annotated at 10 Hz, including 3D bounding boxes, depth, and semantic segmentation, and comprises sensor configurations including LiDAR, six Cameras, five Radars, an IMU/GNSS module, and a BEV Camera, integrated with HD map data. For closed-loop testing, Bench2Drive systematically collects data from 220 routes uniformly distributed across 44 interactive scenarios (such as lane cutting, overtaking, and detours), 23 weather conditions (including clear, foggy, and rainy conditions), and 12 distinct towns in CARLA v2 (encompassing urban, rural, and university environments). 

Furthermore, FUMP evaluates its driving performance on the NAVSIM dataset, a benchmark constructed from OpenScene~\cite{openscene3dscene} for vehicle planning in real-world scenarios. NAVSIM integrates data from eight cameras and five LiDAR sensors to achieve full 360° perception, with annotations provided at 2Hz, including HD maps and object bounding boxes. It employs non-reactive simulation and closed-loop metrics to assess planning performance comprehensively. In this study, we adopt NAVSIM's proposed PDM scores (PDMS)~\cite{navsim}, a weighted composite of five sub-metrics: No At-fault Collisions (NC), Drivable Area Compliance (DAC), Time-to-Collision (TTC), Comfort (Comf.), and Ego Progress (EP).

Notably, NAVSIM does not provide annotated motion data. We derive the motion information through coordinate transformation based on its existing annotations. The detailed procedure is as follows. For each target object with a unique tracking identifier, we extract its pose data $Pose_{t_{i}}^{e} = [x_{t_{i}}^{e},y_{t_{i}}^{e},z_{t_{i}}^{e},yaw_{t_{i}}^{e}], i = 0,..,T$ from the annotation files. And the pose data is collected in the corresponding ego vehicle's coordinate system. We transform the object poses from the ego vehicle's coordinate system to the world coordinate system using the corresponding transformation matrix $\mathbb{T}_{ego \to world}^{t_{i}}$. Based on the object's pose in the world coordinate system at timestamp $t_{0}$, we construct transformation matrix $\mathbb{T}_{world \to target}^{t_{0}}$ to convert coordinates from the world system to the object's local coordinate system as follows. 
\begin{equation}
  \begin{aligned}
   \mathbb{T} = \left[\begin{array}{cccc}
   \cos \psi & \sin \psi & 0 & -x_{t_{0}}^{w} \cos \psi-y_{t_{0}}^{w} \sin \psi \\
   -\sin \psi & \cos \psi & 0 & x_{t_{0}}^{w} \sin \psi-y_{t_{0}}^{w} \cos \psi \\
   0 & 0 & 1 & -z_{t_{0}}^{w} \\
   0 & 0 & 0 & 1
\end{array}\right]
\end{aligned}
\label{matrix}
\end{equation}
Here, $Pose_{t_{0}}^{w} = [x_{t_{0}}^{w},y_{t_{0}}^{w},z_{t_{0}}^{w},\psi]$ represents the target object's position and orientation in the world coordinate system at $t_{0}$. And $\psi = yaw_{t_{0}}^{e} + yaw_{t_{0}}$. $yaw_{t_{0}}$ represents the yaw of ego vehicle in the world coordinate system at $t_{0}$. Therefore, for any given timestamp $t_{i}$, the target object's pose in the ego vehicle's coordinate system $Pose_{t_{i}}^{e}$ can be transformed to its corresponding pose in the $t_{0}$ target object coordinate system through the following rigid transformation,
\begin{equation}
\begin{aligned}
   Pose_{t_{i}}^{t} = \mathbb{T}_{world \to target}^{t_{0}} \cdot \mathbb{T}_{ego \to world}^{t_{i}} \cdot Pose_{t_{i}}^{e}.
\end{aligned}
\label{trans}
\end{equation}

The complete computational pipeline is formally described in Algorithm~\ref{algorithm:TrajPooler}.

\begin{algorithm}[]
\caption{Coordinate Transformation Chain} \label{algorithm:TrajPooler}
\KwIn{
    GT Boxes: $\{Pose_{t_{i}}^{e}\} \in \mathbb{R}^{T \times 4}, $ $i=0,...,T$, \\
    Pose Matrix $\{\mathbb{T}_{ego\to world}^{t_{i}}\} \in \mathbb{R}^{T \times 4 \times 4}, $ $i=0,...,T$,
}
\KwOut{Motion Trajectory $\tau \in \mathbb{R}^{T \times 2}$}

\textbf{Convert gt box to world coordinate system:} $Pose_{t_{i}}^{w} = \mathbb{T}_{ego\to world}^{t_{i}} \cdot Pose_{t_{i}}^{e} $ 

\textbf{Construct Pose Matrix $\mathbb{T}_{world\to target}^{t_{0}}$:} $\mathbb{T}_{world\to target}^{t_{0}} = Construct(Pose_{t_{0}}^{w})$ 

\textbf{Convert gt box to target coordinate system:} $Pose_{t_{i}}^{t} = \mathbb{T}_{world\to target}^{t_{0}} \cdot Pose_{t_{i}}^{w}$ 

\textbf{Get Motion Trajectory $\tau$:} $\tau.append(Pose_{t_{i}}^{t}[0:2])$
\label{al:chain}
\end{algorithm}


\textit{\textbf{For long-tail testing}}, FUMP method offers a distinct advantage by leveraging motion data to augment scarce planning samples, thereby enhancing the performance of end-to-end approaches. To validate this, we curate three long-tail test datasets, NuScenes-LT, Bench2drive-LT and NAVSIM-LT through the following systematic process. We applied the OPTICS adaptive clustering algorithm to ego-trajectories from the validation set, rank the clusters by population size, and select samples from the $k$ smallest clusters to construct our long-tail datasets. The hyperparameter $k$ is set to 35 for NuScenes, 45 for Bench2Drive and 73 for NAVSIM. The NuScenes-LT dataset contains 749 samples and for Bench2Drive-LT dataset, we choose 1218 samples to construct it. Finally, for NAVSIM-LT, we select 1481 driving clips to form it. 

Furthermore, FUMP is evaluated on the ADV-NuScenes~\cite{challenger} dataset to assess its long-tail learning capability in adversarial driving scenarios. Derived from the NuScenes dataset, ADV-NuScenes~\cite{challenger} comprises 6,115 samples across 150 physically plausible yet highly realistic adversarial driving scenarios. The dataset encompasses diverse aggressive driving behaviors, including cut-ins, sudden lane changes, rear-end collisions, and blind-spot intrusions. It is primarily designed for benchmarking collision rates in adversarial environments.


\begin{figure}[t]
    \centering
    \includegraphics[width=\linewidth]{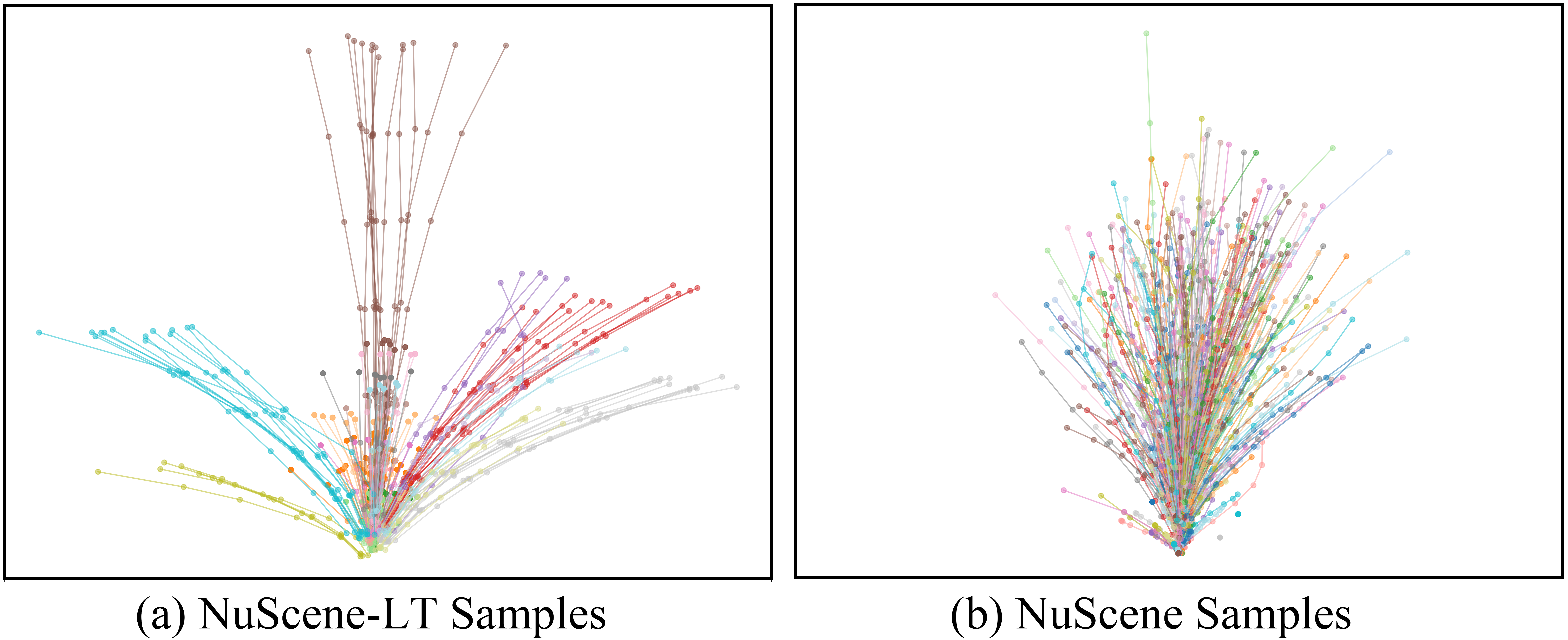}
    \caption{
    \textbf{The visualization of NuScenes-LT trajectories compared to NuScenes} (with NuScenes-LT removed). This demonstrates that NuScenes-LT includes more long-tail scenarios, such as turning maneuvers, than the original dataset.}
    \label{fig:dataset_fs}
\end{figure}

\subsubsection{Evaluation Metrics for Planning}
For open-loop planning evaluation, we employ conventional $L_{2}$ error and CR (collision rate) as metrics, aligning with the computation method of SparseDrive~\cite{sun2024sparsedrive}. Regarding closed-loop planning assessment, in Bench2drive ,we follow the Bench2Drive~\cite{Bench2drive} dataset setting, measuring DS (Driving Score) and SR (Success Rate (\%)). In NAVSIM, we adopt NAVSIM's proposed PDM scores (PDMS)~\cite{navsim}, a weighted composite of five sub-metrics: No At-fault Collisions (NC), Drivable Area Compliance (DAC), Time-to-Collision (TTC), Comfort (Comf.), and Ego Progress (EP).

Additionally, we introduce a novel metric, CEGR (Cross-Vehicle Efficiency Gain Ratio), to assess motion data utilization efficiency.

\begin{equation}
  \begin{aligned}
   CEGR = \frac{(Acc - Acc_{base})}{(Acc_{base})} \cdot (1 - \frac{D_{ego}}{D_{ego + agent}}) \cdot 100 \%\\
  \end{aligned}
\label{plan_2th}
\end{equation}
CEGR can measure a model's ability to transform motion data into learnable data for planning tasks. Here, $Acc$ and $Acc_{base}$ denote the performance of the proposed model and the baseline method, respectively. $D_{ego}$ represents the training data volume for an ego vehicle, while $D_{agent}$ denotes the training data volume for other vehicles.


\subsubsection{Implementation Details}
FUMP adopts SparseDrive~\cite{sun2024sparsedrive}, VAD~\cite{jiang2023vad} and Transfuser~\cite{TransFuser} as its baseline framework while adhering to their established training protocols. Notably, although VAD relies on dense BEV representations for perception, FUMP operates entirely on VAD's sparse perceptual queries without requiring scene-level sparse representations. \textit{\textbf{For NuScenes~\cite{nuscenes}}} dataset, we utilize 750 clips for training and 150 clips for validation. The first stage focuses on training detection and online mapping tasks, employing the learning rate of $3e-4$ for 80 epochs with the batch size of 4. Subsequently, the second stage is dedicated to motion prediction and planning tasks, maintaining the same learning rate and batch size for 10 epochs. \textit{\textbf{For Bench2Drive dataset~\cite{Bench2drive}}}, we conduct experiments using the base-scale configuration, allocating 950 clips for training and 50 clips for validation. To adapt for its distinct sampling frequency, we adjust the training schedule. 
For VAD, FUMP is trained for 6 epochs with a learning rate of 2e-4. \textit{\textbf{And for NAVSIM dataset~\cite{navsim}}}, we adopt TransFuser~\cite{TransFuser} as the baseline model and follow the standard Navtrain split for training. The model is trained for 100 epochs with a batch size of 64.


\begin{table*}[tp]
\centering
  \caption{Planning results on the $\operatorname{NuScenes}$~\cite{nuscenes} validation dataset. $^{\ast}$ denotes results reproduced with the official checkpoint.}
  \renewcommand\arraystretch{1.0}
  \setlength{\tabcolsep}{3.1mm} 
  \begin{tabular}{lclcccc cccc c}
\toprule
\multirow{2}{*}{$\operatorname{Method}$} & \multirow{2}{*}{$\operatorname{Input}$} & \multirow{2}{*}{$\operatorname{Backbone}$} & \multicolumn{4}{c}{$\operatorname{L2\ (m)}\downarrow$} & \multicolumn{4}{c}{$\operatorname{Col.\ Rate\ (\%)}\downarrow$} & \multirow{2}{*}{$\operatorname{FPS}\uparrow$} \\
\cmidrule(lr){4-7} \cmidrule(lr){8-11}
& & & 1s & 2s & 3s & $\operatorname{Avg.}$ & 1s & 2s & 3s & $\operatorname{Avg.}$ \\
\midrule
$\operatorname{IL}$~\cite{IL} & $\operatorname{LiDAR}$ & $\operatorname{VoxelNet}$ & 0.44 & 1.15 & 2.47 & \cellcolor{gray!15}1.35 & 0.08 & 0.27 & 1.95 & \cellcolor{gray!15}0.77 & - \\
$\operatorname{NMP}$~\cite{NMP} & $\operatorname{LiDAR}$ & $\operatorname{VoxelNet}$ & 0.53 & 1.25 & 2.67 & \cellcolor{gray!15}1.48 & 0.04 & 0.12 & 0.87 & \cellcolor{gray!15}0.34  & - \\
$\operatorname{FF}$~\cite{FF} & $\operatorname{LiDAR}$ & $\operatorname{VoxelNet}$ & 0.55 & 1.20 & 2.54 & \cellcolor{gray!15}1.43 & 0.06 & 0.17 & 1.07 & \cellcolor{gray!15}0.43 & - \\
$\operatorname{EO}$~\cite{EO} & $\operatorname{LiDAR}$ & $\operatorname{VoxelNet}$ & 0.67 & 1.36 & 2.78 & \cellcolor{gray!15}1.60 & 0.04 & 0.09 & 0.88 & \cellcolor{gray!15}0.33 & - \\
\midrule
$\operatorname{ST-P3}$~\cite{ST_P3} & $\operatorname{Camera}$ & $\operatorname{ResNet50}$ & 1.33 & 2.11 & 2.90 & \cellcolor{gray!15}2.11 & 0.23 & 0.62 & 1.27 & \cellcolor{gray!15}0.71 & 1.6 (RTX3090)\\
$\operatorname{OccNet}$~\cite{OccNet} & $\operatorname{Camera}$ & $\operatorname{ResNet50}$ & 1.29 & 2.13 & 2.99 & \cellcolor{gray!15}2.14 & 0.21 & 0.59 & 1.37 & \cellcolor{gray!15}0.72 & 2.6 (RTX3090)\\
$\operatorname{UniAD}$~\cite{uniad} & $\operatorname{Camera}$ & $\operatorname{ResNet101}$ & 0.45 & 0.70 & 1.04 & \cellcolor{gray!15}0.73 & 0.62 & 0.58 & 0.63 & \cellcolor{gray!15}0.61 & 1.8 $\operatorname{(A100)}$ \\
$\operatorname{VAD}$~\cite{jiang2023vad} & $\operatorname{Camera}$ & $\operatorname{ResNet50}$ & 0.41 & 0.70 & 1.05 & \cellcolor{gray!15}0.72 & 0.03 & 0.19 & 0.43 & \cellcolor{gray!15}0.21 & - \\
$\operatorname{GenAD}$~\cite{zheng2024genad} & $\operatorname{Camera}$ & $\operatorname{ResNet50}$ & 0.36 & 0.83 & 1.55 & \cellcolor{gray!15}0.91 & 0.06 & 0.23 & 1.00 & \cellcolor{gray!15}0.43 &  6.7 (RTX3090)\\
$\operatorname{MomAD}$~\cite{momad} & $\operatorname{Camera}$ & $\operatorname{ResNet50}$ & 0.31 & 0.57 & 0.91 & \cellcolor{gray!15}0.60 & 0.01 & 0.05 & 0.22 & \cellcolor{gray!15}0.09 &  7.8 (RTX4090)\\
$\operatorname{HE-Drive}$~\cite{HE-Drive} & $\operatorname{Camera}$ & $\operatorname{Llama 3.2V}$ & 0.30 & 0.56 & 0.89 & \cellcolor{gray!15}0.58 & 0.00 & 0.03 & 0.14 & \cellcolor{gray!15}0.06 & 10.0$\operatorname{(RTX4090)}$ \\
$\operatorname{SparseDrive^{\ast}}$~\cite{sun2024sparsedrive} & $\operatorname{Camera}$ & $\operatorname{ResNet50}$ & 0.29 & 0.58 & 0.96 & \cellcolor{gray!15}0.61 & \textbf{0.01} & 0.05 & 0.18 & \cellcolor{gray!15}0.08 & \textbf{9.0 $\operatorname{(RTX4090)}$} \\
\rowcolor{gray!15} $\operatorname{FUMP\ (Ours)}$ & $\operatorname{Camera}$ & $\operatorname{ResNet50}$ & \textbf{0.16} & \textbf{0.38} & \textbf{0.69} & \cellcolor{gray!15}\textbf{0.39} & \textbf{0.01} & \textbf{0.03} & \textbf{0.14} & \cellcolor{gray!15}\textbf{0.06} & 5.9 $\operatorname{(RTX4090)}$ \\
\bottomrule
\end{tabular}
\label{tab_nuscenes_planning}
\end{table*}

\begin{table*}[]
\centering
  \caption{Open-loop and closed-loop results of E2E-AD methods in Bench2Drive~\cite{Bench2drive}. Avg. L2 is averaged over the predictions in 2 seconds under 2Hz. $*$ denotes using expert feature distillation.}
\renewcommand\arraystretch{1.0}
  \setlength{\tabcolsep}{7.2mm}
\begin{tabular}{lcccc}
\toprule
\multicolumn{1}{l|}{\multirow{2}{*}{$\operatorname{Method}$}} & \multicolumn{1}{c|}{\multirow{2}{*}{$\operatorname{Input}$}} &  \multicolumn{1}{c|}{Open-loop Metric} & \multicolumn{2}{c}{Close-loop Metric} \\ \cmidrule{3-5}
                    \multicolumn{1}{c|}{}    & \multicolumn{1}{c|}{}                       & \multicolumn{1}{c|}{$\operatorname{Avg.L2(m)}\downarrow$}  & \multicolumn{1}{c|}{Driving Score$\uparrow$}      & Success Rate$\uparrow$      \\ \midrule
                    \multicolumn{1}{l|}{Think2Twice$^{*}$~\cite{thiktwice}} & \multicolumn{1}{c|}{Ego State + 6 \xspace Cameras} & \multicolumn{1}{c|}{\textbf{0.95}} & 62.44 & 31.23\\
                    \multicolumn{1}{l|}{DriveAdapter$^{*}$~\cite{jia2023driveadapter}} & \multicolumn{1}{c|}{Ego State + 6 \xspace Cameras} & \multicolumn{1}{c|}{1.01} & 64.22 & 33.08\\
                    \multicolumn{1}{l|}{WoTE$^{*}$~\cite{wote}} & \multicolumn{1}{c|}{Ego State + 6 \xspace Cameras} & \multicolumn{1}{c|}{-} & 61.71 & 31.36\\
                    \multicolumn{1}{l|}{DiffAD$^{*}$~\cite{diffad}} & \multicolumn{1}{c|}{Ego State + 6 \xspace Cameras} & \multicolumn{1}{c|}{1.55} & \textbf{67.92} & \textbf{38.64}\\
                    \midrule
                        \multicolumn{1}{l|}{UniAD-Tiny~\cite{uniad}} &     \multicolumn{1}{c|}{Ego State + 6 \xspace Cameras}  &   \multicolumn{1}{c|}{\cellcolor{gray!15} 0.80}                &       \cellcolor{gray!15}40.73            &       \cellcolor{gray!15}13.18       \\
                        \multicolumn{1}{l|}{UniAD-Base~\cite{uniad}} &   \multicolumn{1}{c|}{Ego State + 6 \xspace Cameras} &     \multicolumn{1}{c|}{\cellcolor{gray!15}\textbf{0.73}}             &        \cellcolor{gray!15}\textbf{45.81}            &    \cellcolor{gray!15}16.36               \\
                        \multicolumn{1}{l|}{VAD~\cite{jiang2023vad}} &  \multicolumn{1}{c|}{Ego State + 6 \xspace Cameras} &       \multicolumn{1}{c|}{\cellcolor{gray!15}0.91}           &      \cellcolor{gray!15}42.35              & \cellcolor{gray!15}15.00 \\  \multicolumn{1}{l|}{AD-MLP~\cite{fusionad}} & \multicolumn{1}{c|}{Ego State} &   \multicolumn{1}{c|}{\cellcolor{gray!15}3.64}               &         \cellcolor{gray!15}18.05   & \cellcolor{gray!15}0.00 \\ \multicolumn{1}{l|}{FUMP} &   \multicolumn{1}{c|}{Ego State + 6 \xspace Cameras}& \multicolumn{1}{c|}{\cellcolor{gray!15}0.80}              &     \cellcolor{gray!15}45.67               & \cellcolor{gray!15}\textbf{16.36}  \\ \bottomrule                 
\end{tabular}
\label{tab:bench2drive}
\end{table*}

\begin{table*}[th]
\centering
\caption{Multi-ability results of E2E-AD methods in Bench2Drive dataset. \textbf{Black bold} represents the optimal result.}
\renewcommand\arraystretch{0.75}
\setlength{\tabcolsep}{5.8mm}
\begin{tabular}{lllllll}
\toprule
\multicolumn{1}{l|}{\multirow{2}{*}{Method}} & \multicolumn{6}{c}{Ability(\%)}                                         \\ \cmidrule{2-7} 
\multicolumn{1}{c|}{}                        & Merging & Overtaking & Emergency Brake & Give Way & Traffic Sign & Mean \\ \midrule
\multicolumn{1}{l|}{AD-MLP~\cite{fusionad}} &    \multicolumn{1}{c}{0.00}     &       \multicolumn{1}{c}{0.00}      &         \multicolumn{1}{c}{0.00}         &    \multicolumn{1}{c}{0.00}      &      \multicolumn{1}{c}{4.35}        &     \multicolumn{1}{c}{0.00}  \\
\multicolumn{1}{l|}{UniAD-Tiny~\cite{uniad}}                                   &  \multicolumn{1}{c}{8.89}       &     \multicolumn{1}{c}{9.33}       &       \multicolumn{1}{c}{20.00}          &     \multicolumn{1}{c}{20.00}     &    \multicolumn{1}{c}{15.43}          &    \multicolumn{1}{c}{14.73}  \\
\multicolumn{1}{l|}{UniAD-Base~\cite{uniad}}                                   &    \multicolumn{1}{c}{\textbf{14.10}}     &    \multicolumn{1}{c}{17.78}        &        \multicolumn{1}{c}{\textbf{21.67}}         &     \multicolumn{1}{c}{10.00}     &     \multicolumn{1}{c}{14.21}         &   \multicolumn{1}{c}{15.55}   \\
\midrule
\multicolumn{1}{l|}{VAD~\cite{jiang2023vad}}                                          &    \multicolumn{1}{c}{8.11}     &      \multicolumn{1}{c}{\textbf{24.44}}      &        \multicolumn{1}{c}{18.64}         &     \multicolumn{1}{c}{20.00}     &       \multicolumn{1}{c}{19.15}       &     \multicolumn{1}{c}{18.07} \\
\multicolumn{1}{l|}{\cellcolor{gray!15}FUMP}                                  &    \multicolumn{1}{c}{\cellcolor{gray!15}\textbf{12.50}}     &      \multicolumn{1}{c}{\cellcolor{gray!15}24.44}      &         \multicolumn{1}{c}{\cellcolor{gray!15}20.00}        &     \multicolumn{1}{c}{\cellcolor{gray!15}\textbf{21.50}}     &      \multicolumn{1}{c}{\cellcolor{gray!15}\textbf{19.15}}        &  \multicolumn{1}{c}{\cellcolor{gray!15}\textbf{19.51}} \\
\bottomrule
\end{tabular}
\label{tab:detailbench2drive}
\end{table*}

\begin{table*}[]
\centering
  \caption{Comparison on planning-oriented NAVSIM navtest split with closed-loop metrics. For a fair comparison, we reproduc Transfuser, 
  which outputs multimodal trajectories and takes clustered plan anchors as the model input.}
\renewcommand\arraystretch{1.0}
\setlength{\tabcolsep}{3.8mm}
\begin{tabular}{lcccccccc}
\toprule
\multicolumn{1}{l|}{Method} & \multicolumn{1}{c}{Input} & \multicolumn{1}{c|}{Img Backbone} & NC  $\uparrow$                & \multicolumn{1}{c}{DAC}$\uparrow$ & TTC $\uparrow$                 & Comf.$\uparrow$                & EP$\uparrow$                  & PDMS $\uparrow$ \\
\midrule
UniAD~\cite{uniad} & Camera & ResNet-34 & 97.8 & 91.9 & 92.9 & 100 & 78.8 &83.4 \\
LTF~\cite{TransFuser} & Camera & ResNet-34 & 97.4 & 92.8 & 92.4 & 100 & 79.0 & 83.8 \\
PARA-Drive~\cite{paradrive} & Camera & ResNet-34 & 97.9 & 92.4 & 93.0 & 99.8 & 79.3 & 84.0 \\
DRAMA~\cite{drama} & Camera \& LiDAR & ResNet-34 & 98.0 & 93.1 & 94.8 & 100 & 80.1 & 85.5 \\
VADv2~\cite{chen2024vadv2} & Camera \& LiDAR & ResNet-34 & 97.2 & 89.1 & 91.6 & 100 & 76.0 & 80.9 \\
Hydra-MDP~\cite{li2024hydra} & Camera \& LiDAR & ResNet-34 & \textbf{98.3} & 96.0 & 94.6 & 100 & 78.7 & 86.5 \\
DiffusionDrive~\cite{diffusiondrive} & Camera \& LiDAR & ResNet-34 & 98.2 & 96.2 & \textbf{94.7} & 100 & \textbf{82.2} & \textbf{88.1} \\ 
\midrule
Transfuser~\cite{TransFuser} & Camera \& LiDAR & ResNet-34 & 97.7 & 92.8 & 92.8 & 100 & 79.2 & 84.0 \\ 
$\operatorname{Transfuser}^{\ast}$~\cite{TransFuser} & Camera \& LiDAR & ResNet-34 & 96.2 & 95.4 & 90.7 & 100 & 80.7 & 85.1 \\ 
\cellcolor{gray!15} FUMP & \cellcolor{gray!15} Camera \& LiDAR & \cellcolor{gray!15} ResNet-34 & \cellcolor{gray!15}98.1 & \cellcolor{gray!15}\textbf{96.2} & \cellcolor{gray!15}94.2 & \cellcolor{gray!15}\textbf{100} & \cellcolor{gray!15}82.0 & \cellcolor{gray!15}87.8 \\ 
\bottomrule
\end{tabular}
\label{tab_navsim}
\end{table*}

\begin{table*}[th]
\centering
\caption{Planning results on NuScene-LT and Bench2Drive-LT validation datasets. FUMP adopts SparseDrive and VAD as its baselines for NuScene and Bench2Drive, respectively. \textcolor{blue}{Blue} indicates improvement.}
\renewcommand\arraystretch{1.0}
\setlength{\tabcolsep}{2.0mm}
\begin{tabular}{lcccccccccccccccc}
\toprule
\multirow{3}{*}{Method} & \multicolumn{8}{c}{NuScene-LT}                   & \multicolumn{8}{c}{Bench2Drive-LT}               \\ \cmidrule(lr){2-9} \cmidrule(lr){10-17}
                        & \multicolumn{4}{c}{L2 (m) $\downarrow$} & \multicolumn{4}{c}{Col (\%) $\downarrow$} & \multicolumn{4}{c}{L2 (m) $\downarrow$} & \multicolumn{4}{c}{Col (\%) $\downarrow$} \\
                         \cmidrule(lr){2-5} \cmidrule(lr){6-9} \cmidrule(lr){10-13} \cmidrule(lr){14-17} 
                        & 1s  & 2s  & 3s  & avg  & 1s   & 2s  & 3s  & avg  & 1s  & 2s  & 3s  & avg  & 1s   & 2s  & 3s  & avg  \\
\midrule
VAD~\cite{jiang2023vad} &  0.45   &  0.81   &  2.34   &   1.20   &   0.05   &  0.27   &  1.09   &  0.47    &  0.55  &  0.93  &  1.46  &  0.98   &  0.00  &  0.00  &  0.00  &   0.00  \\
SparseDrive~\cite{sun2024sparsedrive} &  0.29   &  0.59   &  0.99   &   0.63   &   0.00   &  0.00   &  0.29   &   0.09   &  -  &  -  &  -  &  -   &  -  &  -  &  -  &   -  \\
\cellcolor{gray!15}FUMP   &  \cellcolor{gray!15}\textbf{0.15}  &  \cellcolor{gray!15}\textbf{0.39}   &  \cellcolor{gray!15}\textbf{0.80}   &   \cellcolor{gray!15}\textbf{0.45}   &   \cellcolor{gray!15}\textbf{0.00}   &   \cellcolor{gray!15}\textbf{0.00}  &   \cellcolor{gray!15}\textbf{0.09}  &   \cellcolor{gray!15}\textbf{0.03}   &  \cellcolor{gray!15}\textbf{0.50}   &  \cellcolor{gray!15}\textbf{0.78}   &  \cellcolor{gray!15}\textbf{1.33}   &   \cellcolor{gray!15}\textbf{0.87}   &   \cellcolor{gray!15}\textbf{0.00}   &  \cellcolor{gray!15}\textbf{0.00}   &  \cellcolor{gray!15}\textbf{0.00}   & \cellcolor{gray!15}\textbf{0.00} \\
\cellcolor{gray!15}\textit{+improvement}   &  \cellcolor{gray!15}\textit{\textcolor{blue}{-0.14}}  &  \cellcolor{gray!15}\textit{\textcolor{blue}{-0.20}}   &  \cellcolor{gray!15}\textit{\textcolor{blue}{-0.19}}   &   \cellcolor{gray!15}\textit{\textcolor{blue}{-0.18}}   &   \cellcolor{gray!15}\textit{\textcolor{blue}{-0.00}}   &   \cellcolor{gray!15}\textit{\textcolor{blue}{-0.00}}  &   \cellcolor{gray!15}\textit{\textcolor{blue}{-0.20}}  &   \cellcolor{gray!15}\textit{\textcolor{blue}{-0.06}}   &  \cellcolor{gray!15}\textit{\textcolor{blue}{-0.05}}   &  \cellcolor{gray!15}\textit{\textcolor{blue}{-0.15}}   &  \cellcolor{gray!15}\textit{\textcolor{blue}{-0.13}}   &   \cellcolor{gray!15}\textit{\textcolor{blue}{-0.11}}   &   \cellcolor{gray!15}\textit{\textcolor{blue}{-0.00}}   &  \cellcolor{gray!15}\textit{\textcolor{blue}{-0.00}}   &  \cellcolor{gray!15}\textit{\textcolor{blue}{-0.00}}   & \cellcolor{gray!15}\textit{\textcolor{blue}{-0.00}} \\
\bottomrule
\end{tabular}
\label{tab:fs}
\end{table*}

\begin{table*}[]
\centering
  \caption{Planning results on NAVSIM-LT navtest split with closed-loop metrics. FUMP adopts Transfuser as its baseline. \textcolor{blue}{Blue} indicates improvement.}
\renewcommand\arraystretch{1.0}
\setlength{\tabcolsep}{3.7mm}
\begin{tabular}{lcccccccc}
\toprule
\multicolumn{1}{l|}{Method} & \multicolumn{1}{c}{Input} & \multicolumn{1}{c|}{Img Backbone} & NC  $\uparrow$                & \multicolumn{1}{c}{DAC}$\uparrow$ & TTC $\uparrow$                 & Comf.$\uparrow$                & EP$\uparrow$                  & PDMS $\uparrow$ \\
\midrule
DiffusionDrive~\cite{diffusiondrive} & Camera \& LiDAR & ResNet-34 & 99.6 & 96.5 & \textbf{98.5} & 100 & \textbf{85.3} & \textbf{84.9} \\ 
$\operatorname{Transfuser}^{\ast}$~\cite{TransFuser} & Camera \& LiDAR & ResNet-34 & 99.2 & 95.3 & 97.7 & 100 & 78.8 & 78.0 \\ 
\cellcolor{gray!15} FUMP & \cellcolor{gray!15} Camera \& LiDAR & \cellcolor{gray!15} ResNet-34 & \cellcolor{gray!15}\textbf{99.6} & \cellcolor{gray!15}\textbf{96.6} & \cellcolor{gray!15}98.2 & \cellcolor{gray!15}\textbf{100} & \cellcolor{gray!15}83.7 & \cellcolor{gray!15}83.4 \\ 
\cellcolor{gray!15}\textit{+improvement} & \cellcolor{gray!15}- & \cellcolor{gray!15}- & \cellcolor{gray!15} \textcolor{blue}{0.40}& \cellcolor{gray!15} \textcolor{blue}{1.30}& \cellcolor{gray!15} \textcolor{blue}{0.50}& \cellcolor{gray!15}\textcolor{blue}{0.00}& \cellcolor{gray!15}\textcolor{blue}{4.90}& \cellcolor{gray!15}\textcolor{blue}{5.40}\\
\bottomrule
\end{tabular}
\label{tab_navsim_fs}
\end{table*}

\begin{table}[]
\renewcommand\arraystretch{1.0}
\setlength{\tabcolsep}{2.6mm}
\centering
  \caption{Performance on ADV-NuScenes dataset. We report the collision rate over 3 seconds and their average.}
\begin{tabular}{lcccc}
\toprule
\multicolumn{1}{l|}{Method} & \multicolumn{1}{c}{1s} & 2s                   & 3s                   & avg                  \\
\midrule
UniAD~\cite{uniad}                       & \multicolumn{1}{c}{0.800\%}   & \multicolumn{1}{c}{4.100\%} & \multicolumn{1}{c}{6.960\%} & \multicolumn{1}{c}{3.950\%} \\
VAD~\cite{jiang2023vad}  &   \multicolumn{1}{c}{4.460\%}   &         \multicolumn{1}{c}{7.590\%}             &         \multicolumn{1}{c}{9.080\%}             &          \multicolumn{1}{c}{7.05\%}             \\
SparseDrive~\cite{sun2024sparsedrive}                 &           \multicolumn{1}{c}{\textbf{0.029\%}}             & \multicolumn{1}{c}{0.618\%} &          \multicolumn{1}{c}{2.430\%}            & \multicolumn{1}{c}{ 1.026\%} \\
DiffusionDrive~\cite{diffusiondrive}              &          \multicolumn{1}{c}{0.068\%}              & \multicolumn{1}{c}{1.299\%} & \multicolumn{1}{c}{3.646\%} & \multicolumn{1}{c}{1.671\%} \\
\midrule
\cellcolor{gray!15} FUMP       &          \cellcolor{gray!15} 0.174\%       & \cellcolor{gray!15} \textbf{0.434\%}  &    \cellcolor{gray!15} \textbf{1.570\%}       & \cellcolor{gray!15} \textbf{0.726\%} \\
\bottomrule
\end{tabular}
\label{tab_adv}
\end{table}

\begin{table*}[]
\centering
\caption{Impact of the different modules in FUMP on NuScenes~\cite{nuscenes} and Bench2Drive~\cite{Bench2drive} validation datasets. UMP means FUMP without ECSA and UTTD. ECSA denotes Equivariant Context-Sharing Scene Adapter module, UTTD denotes Unified Two-Stage Trajectory Decoder. \textcolor{blue}{Blue} represents positive improvement, and \textcolor{red}{Red} represents negative improvement. }
\renewcommand\arraystretch{1.0}
\resizebox{\linewidth}{!}{
\begin{tabular}{ccccccccccccccccccccc}
\toprule
\multicolumn{1}{c|}{\multirow{3}{*}{UMP}} & \multicolumn{1}{c|}{\multirow{3}{*}{ECSA}} & \multicolumn{1}{c|}{\multirow{3}{*}{UTTD}} & \multicolumn{9}{c|}{NuScene}                                                                                                                                                                                                       & \multicolumn{9}{c}{Bench2Drive}                                                                                                                                                                                                   \\ \cmidrule(lr){4-12} \cmidrule(lr){13-21}
\multicolumn{1}{c|}{}                     & \multicolumn{1}{c|}{}                      & \multicolumn{1}{c|}{}                      & \multicolumn{3}{c|}{L2(m) $\downarrow$}                                                 & \multicolumn{3}{c|}{Col.Rate(\%) $\downarrow$}                                          & \multicolumn{3}{c|}{CEGR(L2) $\uparrow$}                                              & \multicolumn{3}{c|}{L2(m) $\downarrow$}                                                 & \multicolumn{3}{c|}{Col.Rate(\%) $\downarrow$}                                          & \multicolumn{3}{c}{CEGR(L2) $\uparrow$}                                              \\ \cmidrule(lr){4-6} \cmidrule(lr){7-9} \cmidrule(lr){10-12} \cmidrule(lr){13-15} \cmidrule(lr){16-18} \cmidrule(lr){19-21}
\multicolumn{1}{c|}{}                     & \multicolumn{1}{c|}{}                      & \multicolumn{1}{c|}{}                      & \multicolumn{1}{c}{2s} & \multicolumn{1}{c}{3s} & \multicolumn{1}{c}{Avg} & \multicolumn{1}{c}{2s} & \multicolumn{1}{c}{3s} & \multicolumn{1}{c}{Avg} & \multicolumn{1}{c}{2s} & \multicolumn{1}{c}{3s} & \multicolumn{1}{c}{Avg} & \multicolumn{1}{c}{2s} & \multicolumn{1}{c}{3s} & \multicolumn{1}{c}{Avg} & \multicolumn{1}{c}{2s} & \multicolumn{1}{c}{3s} & \multicolumn{1}{c}{Avg} & \multicolumn{1}{c}{2s} & \multicolumn{1}{c}{3s} & \multicolumn{1}{c}{Avg} \\
\midrule
 & & & 0.58 & 0.96 & 0.61 & 0.05 & 0.18 & 0.08 & 0.00 & 0.00 & 0.00 & 0.88 & 1.32 & 0.91 & 0.00 & 0.00 & 0.00 & 0.00 & 0.00 & 0.00 \\
\checkmark  & & & 0.56 & 0.91 & 0.58 & 0.05 & 0.20 & 0.09 & \textcolor{blue}{1.47} & \textcolor{blue}{2.22} & \textcolor{blue}{2.09} & 0.89 & 1.28 & 0.90 & 0.00 & 0.00 & 0.00 & \textcolor{red}{-0.48} & \textcolor{blue}{1.29} & \textcolor{blue}{0.46} \\
 \checkmark & \checkmark & & 0.51 & 0.84 & 0.54 & 0.05 & 0.22 & 0.09 & 
\textcolor{blue}{5.14} & \textcolor{blue}{5.33} & \textcolor{blue}{4.89} & 0.86 & 1.29 & 0.89 & 0.00 & 0.00 & 0.00 & \textcolor{blue}{0.96} & \textcolor{blue}{0.96} & \textcolor{blue}{0.93} \\
 \checkmark & \checkmark & \checkmark &  0.38 & 0.69 & 0.39 & 0.03 & 0.14 & 0.06 & \textcolor{blue}{14.70} & \textcolor{blue}{11.99} & \textcolor{blue}{14.68} & 0.68 & 1.22 & 0.80 & 0.00 & 0.00 & 0.00 & \textcolor{blue}{9.70} & \textcolor{blue}{3.23} & \textcolor{blue}{5.15} \\
\bottomrule
\end{tabular}
\label{tab:ablation}
}
\end{table*}

\begin{table*}[]
\centering
\caption{Impact of the different modules in FUMP on NuScenes-LT~\cite{nuscenes} and Bench2Drive-LT~\cite{Bench2drive} validation datasets. }
\renewcommand\arraystretch{1.0}
\resizebox{\linewidth}{!}{
\begin{tabular}{ccccccccccccccccccccc}
\toprule
\multicolumn{1}{c|}{\multirow{3}{*}{UMP}} & \multicolumn{1}{c|}{\multirow{3}{*}{ECSA}} & \multicolumn{1}{c|}{\multirow{3}{*}{UTTD}} & \multicolumn{9}{c|}{NuScene-LT}                                                                                                                                                                                                       & \multicolumn{9}{c}{Bench2Drive-LT}                                                                                                                                                                                                   \\ \cmidrule(lr){4-12} \cmidrule(lr){13-21}
\multicolumn{1}{c|}{}                     & \multicolumn{1}{c|}{}                      & \multicolumn{1}{c|}{}                      & \multicolumn{3}{c|}{L2(m)}                                                 & \multicolumn{3}{c|}{Col.Rate(\%)}                                          & \multicolumn{3}{c|}{CEGR(L2)}                                              & \multicolumn{3}{c|}{L2(m)}                                                 & \multicolumn{3}{c|}{Col.Rate(\%)}                                          & \multicolumn{3}{c}{CEGR(L2)}                                              \\ \cmidrule(lr){4-6} \cmidrule(lr){7-9} \cmidrule(lr){10-12} \cmidrule(lr){13-15} \cmidrule(lr){16-18} \cmidrule(lr){19-21}
\multicolumn{1}{c|}{}                     & \multicolumn{1}{c|}{}                      & \multicolumn{1}{c|}{}                      & \multicolumn{1}{c}{2s} & \multicolumn{1}{c}{3s} & \multicolumn{1}{c}{Avg} & \multicolumn{1}{c}{2s} & \multicolumn{1}{c}{3s} & \multicolumn{1}{c}{Avg} & \multicolumn{1}{c}{2s} & \multicolumn{1}{c}{3s} & \multicolumn{1}{c}{Avg} & \multicolumn{1}{c}{2s} & \multicolumn{1}{c}{3s} & \multicolumn{1}{c}{Avg} & \multicolumn{1}{c}{2s} & \multicolumn{1}{c}{3s} & \multicolumn{1}{c}{Avg} & \multicolumn{1}{c}{2s} & \multicolumn{1}{c}{3s} & \multicolumn{1}{c}{Avg} \\
\midrule
 & & & 0.59 & 0.99 & 0.63 & 0.00 & 0.29 & 0.09 & 0.00 & 0.00 & 0.00 & 0.93 & 1.46 & 0.98 & 0.00 & 0.00 & 0.00 & 0.00 & 0.00 & 0.00 \\
\checkmark  & & & 0.59 & 0.96 & 0.62 & 0.00 & 0.29 & 0.09 & \textcolor{red}{-0.00} & \textcolor{blue}{1.29} & \textcolor{blue}{0.67} & 0.91 & 1.45 & 0.97 & 0.00 & 0.00 & 0.00 & \textcolor{blue}{0.92} & \textcolor{blue}{0.29} & \textcolor{blue}{0.43} \\
 \checkmark & \checkmark & & 0.57 & 0.94 & 0.60 & 0.00 & 0.29 & 0.09 & \textcolor{blue}{1.44} & \textcolor{blue}{2.15} & \textcolor{blue}{2.03} & 0.91 & 1.43 & 0.95 & 0.00 & 0.00 & 0.00 & \textcolor{blue}{0.92} & \textcolor{blue}{0.87} & \textcolor{blue}{1.30}\\
 \checkmark & \checkmark & \checkmark & 0.39 & 0.80 & 0.45 & 0.00 & 0.09 & 0.03 & \textcolor{blue}{14.40} & \textcolor{blue}{8.18} & \textcolor{blue}{12.18} & 0.78 & 1.33 & 0.87 & 0.00 & 0.00 & 0.00 & \textcolor{blue}{5.04} & \textcolor{blue}{4.38} & \textcolor{blue}{5.65} \\
\bottomrule
\end{tabular}
\label{tab:ablation_fs}
}
\end{table*}

\begin{table}[]
\centering
\caption{Impact of the different modules in FUMP on Bench2Drive Close-Loop~\cite{Bench2drive} validation dataset.}
\resizebox{\linewidth}{!}{
\begin{tabular}{ccccccc}
\toprule
\multicolumn{1}{c}{UMP} & \multicolumn{1}{c}{ECSA} & \multicolumn{1}{c}{UTTD} & \multicolumn{1}{c}{Drive Score} & \multicolumn{1}{c}{Success Rate} & \multicolumn{1}{c}{CEGR (DS)} & \multicolumn{1}{c}{CEGR (SR)} \\
\midrule
 &   &    & 42.35 & 15.00 & 0.00 & 0.00 \\
 \checkmark &   &    & 38.87 & 13.63 &  \textcolor{red}{-3.50}  &  \textcolor{red}{-3.89} \\
 \checkmark & \checkmark &   &  42.88  & 15.00 & \textcolor{blue}{0.53}  & \textcolor{blue}{0.00} \\
 \checkmark & \checkmark & \checkmark &  45.67  &  16.36 &  \textcolor{blue}{3.34}  &  \textcolor{blue}{3.86} \\
 \bottomrule
\end{tabular}
\label{tab:ablation_close}
}
\end{table}

\begin{table}[]
\renewcommand\arraystretch{0.75}
\setlength{\tabcolsep}{1.7mm}
\centering
\caption{Ablation studies of the impact of the different modules in FUMP on NAVSIM~\cite{navsim} navtest split dataset.}
\begin{tabular}{ccccccccc}
\toprule
UMP & ECSA & UTTD & NC & DAC & TTC & Comf. & EP & PDMS \\
\midrule
   &      &      &  96.2  &  95.4   &  90.7   & 100 & 80.7 & 85.1 \\
\checkmark   &      &      &  96.6  &   95.5  &   91.3  & 100 & 81.1 & 85.6 \\
\checkmark   &   \checkmark   &      &  96.9  &   95.6  & 91.8   & 100 & 81.3 & 86.0 \\
\checkmark   &   \checkmark   &   \checkmark   &  98.1  &  96.2  &  94.2  &  100 & 82.0 & 87.8 \\
\bottomrule
\end{tabular}
\label{tab_ablation_nav}
\end{table}

\begin{table}[]
\renewcommand\arraystretch{0.75}
\setlength{\tabcolsep}{1.7mm}
\centering
\caption{Impact of the different modules in FUMP on NAVSIM-LT~\cite{navsim} navtest split dataset.}
\begin{tabular}{ccccccccc}
\toprule
UMP & ECSA & UTTD & NC & DAC & TTC & Comf. & EP & PDMS \\
\midrule
   &      &      &  99.2  &  95.3   &  97.7   & 100 & 78.8 & 78.0 \\
\checkmark   &      &      &  99.1  &   95.8  &   97.7  & 100 & 80.2 & 79.5 \\
\checkmark   &   \checkmark   &      &  99.3  &   96.0  & 97.8   & 100 & 81.3 & 80.6 \\
\checkmark   &   \checkmark   &   \checkmark   &  99.6  &  96.6  &  98.2  &  100 & 83.7 & 83.4 \\
\bottomrule
\end{tabular}
\label{tab_ablation_nav_fs}
\end{table}

\begin{table}[]
\renewcommand\arraystretch{0.75}
\setlength{\tabcolsep}{1.0mm}
\centering
  \caption{Impact of the different modules in FUMP on ADV-NuScenes validation datasets. C.R. represents Col.Rate(\%).}
\begin{tabular}{cccccccc}
\toprule
\multirow{2}{*}{UMP} & \multicolumn{1}{c}{\multirow{2}{*}{ECSA}} & \multicolumn{1}{c|}{\multirow{2}{*}{UTTD}} & \multicolumn{4}{c|}{Col.Rate(\%)} & \multicolumn{1}{c}{CEGR(C.R)}  \\ \cmidrule{4-7} \cmidrule{8-8}
& \multicolumn{1}{c}{}  & \multicolumn{1}{c|}{} & \multicolumn{1}{c}{1s} & \multicolumn{1}{c}{2s} & \multicolumn{1}{c}{3s} & \multicolumn{1}{c|}{avg} & \multicolumn{1}{c}{avg} \\
\midrule
 &   &   & 0.029\% & 0.618\% & 2.430\% & 1.026\% & 0.00 \\
 \checkmark &   &   & 0.357\%  & 0.758\%  &  2.050\% & 1.055\% & \textcolor{red}{-1.25} \\
 \checkmark &  \checkmark &  & 0.319\% & 0.711\%  &  1.922\% &  0.984\% & \textcolor{blue}{2.86} \\
 \cellcolor{gray!15} \checkmark & \cellcolor{gray!15} \checkmark &  \cellcolor{gray!15} \checkmark &  \cellcolor{gray!15} 0.174\% & \cellcolor{gray!15} 0.434\%  &  \cellcolor{gray!15} 1.570\% & \cellcolor{gray!15} 0.726\% &  \cellcolor{gray!15} \textcolor{blue}{12.5} \\
\bottomrule
\end{tabular}
\label{tab_ablation_adv}
\end{table}

\begin{table}[]
\centering
\caption{Impact of the hyper-parameter $Length$ on NuScenes-LT, Bench2Drive-LT and NAVSIM-LT Datasets.}
\renewcommand\arraystretch{0.75}
\setlength{\tabcolsep}{2.8mm}
\begin{tabular}{cccccc}
\toprule
\multicolumn{1}{l|}{\multirow{2}{*}{Length}} & \multicolumn{2}{c|}{NuScenes-LT} & \multicolumn{2}{c|}{Bench2Drive-LT} & NAVSIM-LT \\ \cmidrule{2-6}
\multicolumn{1}{c|}{}  & \multicolumn{1}{c|}{L2} & \multicolumn{1}{c|}{Col.Rate} & \multicolumn{1}{c|}{L2} & \multicolumn{1}{c|}{Col.Rate} & PDMS \\
\midrule
100  & 0.49 & 0.04 & 0.91 & 0.00 & 82.6 \\
300  & 0.47 & 0.03 & 0.90 & 0.00 & 82.9 \\
500  & 0.45 & 0.03 & 0.87 & 0.00 & 83.3 \\
700  & 0.45 & 0.03 & 0.87 & 0.00 & 83.4 \\
\bottomrule      
\end{tabular}
\label{tab_length}
\end{table}

\begin{table}[]
\centering
\caption{Impact of the hyper-parameter $\gamma$ on NuScenes-LT, Bench2Drive-LT and NAVSIM-LT Datasets.}
\renewcommand\arraystretch{0.75}
\setlength{\tabcolsep}{3mm}
\begin{tabular}{cccccc}
\toprule
\multicolumn{1}{l|}{\multirow{2}{*}{$\gamma$}} & \multicolumn{2}{c|}{NuScenes-LT} & \multicolumn{2}{c|}{Bench2Drive-LT} & NAVSIM-LT \\ \cmidrule{2-6}
\multicolumn{1}{c|}{} & \multicolumn{1}{c|}{L2} & \multicolumn{1}{c|}{Col.Rate} & \multicolumn{1}{c|}{L2} & \multicolumn{1}{c|}{Col.Rate} & PDMS \\
\midrule
0.1  & 0.47 & 0.03 & 0.87 & 0.00 & 83.4 \\
0.2  & 0.45 & 0.03 & 0.87 & 0.00 & 83.4 \\
0.3  & 0.46 & 0.03 & 0.88 & 0.00 & 83.3 \\
0.4  & 0.46 & 0.03 & 0.88 & 0.00 & 83.1 \\
\bottomrule      
\end{tabular}
\label{tab_gamma}
\end{table}

\begin{table*}[]
\centering
\caption{The robustness of FUMP compared with the baseline SparseDrive on NuScenes validation set. The noise injection methodology is derived from~\cite{benchmark_robustness}, RCE represents the relative performance degradation ratio of the model under noisy conditions.}
\renewcommand\arraystretch{0.75}
\setlength{\tabcolsep}{2.3mm}
\begin{tabular}{lcccccccccc}
\toprule
\multicolumn{1}{l|}{\multirow{3}{*}{Method}} & \multicolumn{10}{c}{Noise}                                                                                                                                          \\ \cmidrule{2-11}
\multicolumn{1}{l|}{}                        & \multicolumn{2}{c|}{Motion Blur} & \multicolumn{2}{c|}{Guassian}                & \multicolumn{2}{c|}{Shear} & \multicolumn{2}{c|}{Scale} & \multicolumn{2}{c}{Uniform} \\ \cmidrule{2-3} \cmidrule{4-5} \cmidrule{6-7} \cmidrule{8-9} \cmidrule{10-11}
\multicolumn{1}{c|}{}                        &   L2(m)   & Col.Rate(\%)    &    L2(m)                  & Col.Rate(\%) &      L2(m)       & Col.Rate(\%)  &      L2(m)      &     Col.Rate(\%)       &        L2(m)     &     Col.Rate(\%)      \\
\midrule
SparseDrive                                  &   0.69   &         0.13                &          0.68            &         0.13             &     0.73                  &  0.18   &      0.77       &      0.16       &       0.69        &      0.16        \\
RCE(\%)                                      &   \textcolor{red}{13.11}   &            \textcolor{red}{62.50}             & \textbf{\textcolor{blue}{11.47}} & \textbf{\textcolor{blue}{38.46}}  & \textcolor{red}{19.67}  &  \textcolor{red}{112.5} &     \textcolor{red}{26.22}       &      \textcolor{red}{100.00}       &  \textbf{\textcolor{blue}{13.11}}       &  \textcolor{red}{100.00} \\
\midrule
FUMP  &  0.42   &  0.09   &   0.44    &   0.09    &   0.40   &   0.09   &  0.40      &    0.08   & 0.46 & 0.08 \\
RCE(\%) &   \textbf{\textcolor{blue}{7.69}}   &  \textbf{\textcolor{blue}{50.00}}  &   \textcolor{red}{12.82}   &  \textcolor{red}{50.00}  &   \textbf{\textcolor{blue}{2.56}}  &   \textbf{\textcolor{blue}{50.00}}   &   \textbf{\textcolor{blue}{2.56}}   &    \textbf{\textcolor{blue}{33.33}}    &  \textcolor{red}{17.90}   &  \textbf{\textcolor{blue}{33.33}} \\
\bottomrule
\end{tabular}
\label{noise_nus}
\end{table*}

\begin{table*}[]
\centering
\caption{The robustness of FUMP compared with the baseline VAD on BenchDrive validation set.}
\renewcommand\arraystretch{0.75}
\setlength{\tabcolsep}{2.7mm}
\begin{tabular}{lcccccccccc}
\toprule
\multicolumn{1}{l|}{\multirow{3}{*}{Method}} & \multicolumn{10}{c}{Noise}                                                                                                                                          \\ \cmidrule{2-11}
\multicolumn{1}{l|}{}                        & \multicolumn{2}{c|}{Motion Blur} & \multicolumn{2}{c|}{Guassian}                & \multicolumn{2}{c|}{Shear} & \multicolumn{2}{c|}{Scale} & \multicolumn{2}{c}{Uniform} \\ \cmidrule{2-3} \cmidrule{4-5} \cmidrule{6-7} \cmidrule{8-9} \cmidrule{10-11}
\multicolumn{1}{c|}{}                        &   L2(m)   & Col.Rate(\%)    &    L2(m)                  & Col.Rate(\%) &      L2(m)       & Col.Rate(\%)  &      L2(m)      &     Col.Rate(\%)       &        L2(m)     &     Col.Rate(\%)      \\
\midrule 
VAD                                  &   0.97   &         0.00                &          0.97            &         0.00             &     1.03                  &  0.00   &      1.05       &      0.00       &       1.01        &      0.00        \\
RCE(\%)                                      &   \textbf{\textcolor{blue}{6.59}}   &            \textcolor{red}{0.00}             & \textcolor{red}{6.59} & \textcolor{red}{0.00}  & \textcolor{red}{13.18}  &  \textcolor{red}{0.00} &     \textcolor{red}{15.38}       &      \textcolor{red}{0.00}       &  \textcolor{red}{10.79}       &  \textcolor{red}{0.00} \\
\midrule
FUMP  &  0.87   &  0.00   &   0.85    &   0.00    &   0.84   &   0.00   &  0.85      &    0.00   & 0.88 & 0.00 \\
RCE(\%) &   \textcolor{red}{8.75}   &  \textcolor{red}{0.00}  &   \textbf{\textcolor{blue}{6.25}}   &  \textcolor{red}{0.00}  &   \textbf{\textcolor{blue}{5.00}}  &   \textcolor{red}{0.00}   &   \textbf{\textcolor{blue}{6.25}}   &   \textcolor{red}{0.00}    &  \textcolor{blue}{10.00}   &  \textcolor{red}{0.00} \\
\bottomrule
\end{tabular}
\label{noise_ben}
\end{table*}

\begin{table}[]
\centering
\caption{The robustness of FUMP compared with the baseline Transfuser on NAVSIM navtest split.}
\renewcommand\arraystretch{0.75}
\setlength{\tabcolsep}{1.1mm}
\begin{tabular}{lccccccc}
\toprule
\multicolumn{1}{c}{Method}     & Noise                        & NC                   & DAC & \multicolumn{1}{c}{TTC} & Comf.                & EP                   & PDMS                 \\
\midrule
\multicolumn{1}{l|}{Transfuser} & \multicolumn{1}{c|}{\multirow{4}{*}{Motion Blur}} & \multicolumn{1}{c}{88.6} &   84.6  & \multicolumn{1}{c}{78.1}    & \multicolumn{1}{c}{99.9} & \multicolumn{1}{c}{62.8} & \multicolumn{1}{c}{65.8} \\
\multicolumn{1}{l|}{RCE(\%)} & \multicolumn{1}{c|}{} &  \textbf{\textcolor{blue}{7.90}}  &  \textcolor{red}{11.32}   &  \textbf{\textcolor{blue}{13.89}}  &      \textbf{\textcolor{blue}{0.00}}                &         \textcolor{red}{21.59}             &     \textcolor{red}{22.76}                 \\
\multicolumn{1}{l|}{FUMP} & \multicolumn{1}{c|}{} &  88.2  &   89.9  &  77.9  &            100          &           66.4           &      69.8                \\
\multicolumn{1}{l|}{RCE(\%)} & \multicolumn{1}{c|}{} &  \textcolor{red}{10.09}  &  \textbf{\textcolor{blue}{6.54}}   &  \textcolor{red}{17.32}  &        \textbf{\textcolor{blue}{0.00}}              &           \textbf{\textcolor{blue}{19.02}}           &          \textbf{\textcolor{blue}{20.05}}            \\
\midrule
\multicolumn{1}{l|}{Transfuser} & \multicolumn{1}{c|}{\multirow{4}{*}{Gaussain}} & \multicolumn{1}{c}{83.3} &  80.7   & \multicolumn{1}{c}{72.2}    & \multicolumn{1}{c}{99.9} & \multicolumn{1}{c}{52.1} & \multicolumn{1}{c}{56.3} \\
\multicolumn{1}{l|}{RCE(\%)} & \multicolumn{1}{c|}{} &  \textcolor{red}{13.40}  &   \textcolor{red}{15.40}  &  \textcolor{red}{20.39}  &     \textbf{\textcolor{blue}{0.00}}   &   \textcolor{red}{35.43}   &         \textcolor{red}{33.92}             \\
\multicolumn{1}{l|}{FUMP} & \multicolumn{1}{c|}{} &  85.9  &   85.1  &  74.7  &          99.9            &          63.3            &     64.3                 \\
\multicolumn{1}{l|}{RCE(\%)} & \multicolumn{1}{c|}{} &  \textbf{\textcolor{blue}{12.43}}  &   \textbf{\textcolor{blue}{11.53}}  &  \textbf{\textcolor{blue}{20.70}}  &  \textbf{\textcolor{blue}{0.00}}   &            \textbf{\textcolor{blue}{22.80}}          &          \textbf{\textcolor{blue}{26.76}}            \\
\midrule
\multicolumn{1}{l|}{Transfuser} & \multicolumn{1}{c|}{\multirow{4}{*}{Shear}} & \multicolumn{1}{c}{93.2} &   86.3  & \multicolumn{1}{c}{84.6}    & \multicolumn{1}{c}{99.9} & \multicolumn{1}{c}{68.9} & \multicolumn{1}{c}{72.3} \\
\multicolumn{1}{l|}{RCE(\%)} & \multicolumn{1}{c|}{} &  \textbf{\textcolor{blue}{3.11}}  &   \textcolor{red}{9.53}  &  \textcolor{blue}{\textbf{6.72}}  &    \textbf{\textcolor{blue}{0.00}}                  &          \textcolor{red}{14.62}            &            \textcolor{red}{15.14}          \\
\multicolumn{1}{l|}{FUMP} & \multicolumn{1}{c|}{} &  92.8  &  92.2  & 84.6   &           99.9           &           74.1           &      77.5                \\
\multicolumn{1}{l|}{RCE(\%)} & \multicolumn{1}{c|}{} &  \textcolor{red}{5.40}  &   \textbf{\textcolor{blue}{4.15}}  &  \textcolor{red}{10.19}  &        \textbf{\textcolor{blue}{0.00}}              &          \textbf{\textcolor{blue}{9.63}}            &          \textbf{\textcolor{blue}{11.73}}            \\
\midrule
\multicolumn{1}{l|}{Transfuser} & \multicolumn{1}{c|}{\multirow{4}{*}{Scale}} & \multicolumn{1}{c}{91.1} &   87.3  & \multicolumn{1}{c}{81.5}    & \multicolumn{1}{c}{99.9} & \multicolumn{1}{c}{64.3} & \multicolumn{1}{c}{69.3} \\
\multicolumn{1}{l|}{RCE(\%)} & \multicolumn{1}{c|}{} &  \textcolor{red}{5.30}  &  \textcolor{red}{8.49}   &  \textcolor{red}{10.14}  &        \textbf{\textcolor{blue}{0.00}}       &  \textcolor{red}{20.32}  &                \textcolor{red}{18.66}      \\
\multicolumn{1}{l|}{FUMP} & \multicolumn{1}{c|}{} &  94.7  &   94.1  &  87.7  &           100           &           78.4           &     81.9                 \\
\multicolumn{1}{l|}{RCE(\%)} & \multicolumn{1}{c|}{} &  \textbf{\textcolor{blue}{3.46}}  &   \textbf{\textcolor{blue}{2.18}}  &  \textbf{\textcolor{blue}{6.90}}  &  \textbf{\textcolor{blue}{0.00}}  &  \textbf{\textcolor{blue}{4.59}}  &   \textbf{\textcolor{blue}{6.71}}  \\
\midrule
\multicolumn{1}{l|}{Transfuser} & \multicolumn{1}{c|}{\multirow{4}{*}{Uniform}} & \multicolumn{1}{c}{94.1} &   88.0  & \multicolumn{1}{c}{86.1}    & \multicolumn{1}{c}{99.9} & \multicolumn{1}{c}{69.2} & \multicolumn{1}{c}{74.3} \\
\multicolumn{1}{l|}{RCE(\%)} & \multicolumn{1}{c|}{} &  \textbf{\textcolor{blue}{2.17}}  &   \textcolor{red}{7.75}  &  \textbf{\textcolor{blue}{5.07}}  &         \textbf{\textcolor{blue}{0.00}}             &        \textcolor{red}{14.25}              &          \textbf{\textcolor{blue}{12.79}}            \\
\multicolumn{1}{l|}{FUMP} & \multicolumn{1}{c|}{} &  88.2  &   92.0  &  77.8  &           99.9           &            71.5          &               72.9       \\
\multicolumn{1}{l|}{RCE(\%)} & \multicolumn{1}{c|}{} &  \textcolor{red}{10.20}  &   \textbf{\textcolor{blue}{4.36}}  &  \textcolor{red}{17.40}  &                  \textbf{\textcolor{blue}{0.00}}    &          \textbf{\textcolor{blue}{12.80}}            &           \textcolor{red}{16.97}           \\
\bottomrule
\end{tabular}
\label{noise_nav}
\end{table}
\subsection{Main Result}
\subsubsection{Plan Results}
\textit{\textbf{NuScenes (Open-Loop) Experiments}}. Table~\ref{tab_nuscenes_planning} presents a comparative evaluation of various end-to-end methods in terms of $L_{2}$ distance error and collision rate. Our proposed method, FUMP, demonstrates state-of-the-art performance, significantly outperforming both LiDAR-based and Camera-based approaches across all evaluated metrics. \textit{Compared to our baseline method}, SparseDrive~\cite{sun2024sparsedrive}, our FUMP method achieves a substantially lower $L_{2}$ error of 0.39 (m) and a reduced collision rate of 0.06\%. This 36.06\% reduction in $L_{2}$ error demonstrates the effectiveness of our approach in achieving more precise and safer trajectory predictions. Among SOTA camera-based methods like UniAD~\cite{uniad}, VAD~\cite{jiang2023vad} and GenAD~\cite{zheng2024genad}, our FUMP model outperforms them in both accuracy and safety. Besides, despite the inherent advantage of LiDAR in depth perception, FUMP surpasses all LiDAR-based methods (IL, NMP, FF, and EO) in both $L_{2}$ error and collision rate. This highlights the superiority of the FUMP's ECSA module, which effectively compensates for the lack of camera depth information through robust feature extraction.

\textit{\textbf{Bench2Drive (Open-Loop and Close-Loop) Experiments}}. As shown in Table~\ref{tab:bench2drive}, experimental results on Bench2Drive demonstrate that our proposed FUMP method achieves superior performance in both open-loop and closed-loop evaluations compared to the VAD~\cite{jiang2023vad} baseline. In detail, FUMP significantly reduces the average L2 error (open-loop metric) from 0.91m to 0.80m, indicating more accurate trajectory prediction. For closed-loop metrics, FUMP outperforms VAD~\cite{sun2024sparsedrive} with a Driving Score improvement from 42.35 to 45.67 and Success Rate increase from 15.00\% to 16.36\%. It should be emphasized that our method's lack of expert knowledge distillation results in comparatively weaker closed-loop performance relative to Think2Twice~\cite{thiktwice} and related approaches. 

Besides, as shown in Table~\ref{tab:detailbench2drive}, we evaluate the performance of state-of-the-art E2E (end-to-end) autonomous driving methods on the Bench2Drive benchmark, assessing key driving abilities including Merging, Overtaking, Emergency Brake, Give Way, and Traffic Sign recognition. The compared methods include AD-MLP~\cite{fusionad}, UniAD-Tiny~\cite{uniad}, UniAD-Base~\cite{uniad}, VAD~\cite{jiang2023vad}(baseline), and our proposed approach, FUMP. The experimental results demonstrate that FUMP consistently surpasses the baseline across all evaluated metrics, achieving relative improvements of 54.87\% in Merging, 7.29\% in Emergency Brake, and 7.50\% in Give Way. This demonstrates FUMP's scalability across diverse driving scenarios. FUMP captures motion and planning trajectories to construct a more comprehensive training corpus that captures a wider spectrum of driving scenarios, thereby enhancing the model's generalization capability.

The superior performance of FUMP stems from its novel integration of motion-aware unified training, which enables joint optimization across diverse planning tasks. Unlike existing approaches that treat driving subtasks independently, FUMP leverages motion priors to enhance generalization, resulting in a more robust and adaptive autonomous driving system. This unified learning paradigm not only improves task-specific performance but also ensures scalability across varied driving scenarios, underscoring FUMP’s potential as a foundational framework for next-generation autonomous vehicles.


\textit{\textbf{NAVSIM (Close-Loop) Experiments}}. As shown in Table~\ref{tab_navsim}, FUMP achieves SOTA results on the NAVSIM dataset for both Comfort (Comf.) and Driving Area Compliance (DAC) metrics, while remaining competitive across other metrics with DiffusionDrive~\cite{diffusiondrive}. Compared to other leading multimodal approaches (VADv2~\cite{chen2024vadv2}, DRAMA~\cite{drama}, and Hydra-MDP~\cite{li2024hydra}), FUMP exhibits significant superiority. Notably, compared to the TransFuser~\cite{TransFuser} baseline, FUMP improves the Ego Process (EP) metric by 1.3\% while substantially enhancing No at-fault Collisions (NC) (1.9\%), Driving Area Compliance (DAC) (0.8\%), and Time to Collision (TTC) (3.5\%) metrics. These results indicate that FUMP not only reliably navigates to target destinations, but also maintains better regulatory compliance and significantly reduces collision risks. This performance advantage stems from FUMP's integrated learning framework that jointly processes motion and planning trajectories, enabling the planner to assimilate broader driving scenarios and strategies. The synergistic learning paradigm ultimately yields more robust decision-making capabilities.

\textit{\textbf{Bench2Drive-LT and NuScenes-LT (Open-Loop) Experiments}}. Table~\ref{tab:fs} summarizes the planning performance on the NuScene-LT and Bench2Drive-LT validation datasets, evaluating key metrics: L2 error and collision rate. We compare FUMP with SparseDrive (baseline), VAD (baseline) and UniAD (prior state-of-the-art). The results demonstrate FUMP’s effectiveness in long-tail driving scenarios, particularly in balancing prediction precision and safety. \textit{In NuScene-LT}, FUMP achieves significant improvements over baseline. It reduces the average L2 error to 0.80 in 3s, outperforming SparseDrive (0.99, 19\% reduction) and VAD (1.20, 33\% reduction). Notably, FUMP also lowers the collision rate to 0.03, a 3× and 15.7× improvement over SparseDrive (0.09) and VAD (0.47), respectively. These gains arise from FUMP’s integration of motion data during training, which alleviates data scarcity in long-tail planning. \textit{In Bench2Drive-LT}, 
FUMP achieves a 0.11 reduction in L2 error compared to the baseline VAD, resulting in an 11.22\% improvement for the baseline. Overall, FUMP consistently outperforms other methods by effectively utilizing motion data for long-tail planning. Its joint optimization of motion and planning tasks improves both accuracy and safety, especially in NuScenes-LT and Bench2Drive-LT.

\textit{\textbf{NAVSIM-LT (Close-Loop) Experiments}}. Based on the results presented in the Table~\ref{tab_navsim_fs}, we conduct a comprehensive comparative analysis of FUMP against SOTA methods and baselines (DiffusionDrive and TransFuser). The results demonstrate that FUMP achieves competitive, even SOTA performance across multiple key metrics. Specifically, FUMP matches DiffusionDrive’s scores in NC (99.6\%) and TTC (98.2\%). Notably, our method attains a 96.6\% DAC score, outperforming DiffusionDrive by 0.1 percentage points, highlighting its superior path compliance. Most significantly, FUMP shows substantial improvements over the TransFuser baseline across all metrics, with particularly remarkable gains in the EP metric. These results conclusively validate the effectiveness of our unified motion-aware learning framework in enhancing both the generalization capability and robustness of driving planning systems.


\textit{\textbf{ADV-NuScenes (Open-Loop) Experiments}}. As highlighted in Table~\ref{tab_adv}, we primarily evaluate collision-related metrics on the challenging Adv-NuScenes benchmark, which features extensive adversarial vehicle behaviors (e.g., lane changes, cut-in maneuvers). FUMP demonstrates exceptional robustness in these scenarios, achieving a 29.2\% reduction in collision rate compared to prior state-of-the-art methods—the most significant improvement recorded on this benchmark. This performance advantage stems from FUMP’s effective learning of motion trajectories patterns, which enables reliable reproduction of adversarial scenarios in novel driving environments. The results underscore FUMP’s scalability and robustness when confronted with real-world adversarial driving conditions.

\subsection{Ablation Study}

\subsubsection{The roles of different modules in FUMP}

\textbf{On the NuScenes and Bench2Drive Datasets (Open-Loop)}. Table~\ref{tab:ablation} presents ablation studies of the FUMP model on NuScenes and Bench2Drive datasets, analyzing the impact of different modules (UMP, ESCA, and UTTD) on performance metrics including L2 (m), Col. Rate, and CEGR (L2), where {UMP} (unified motion planning without ECSA and UTTD) module streamlines model architecture by unifying motion and planning tasks. 
The UMP's limited improvements suggest that a simple, unified task design cannot fully exploit the benefits of motion data for planning tasks. With the introduction of the \textbf{ESCA} module, FUMP demonstrates significant improvements on both the Bench2Drive and NuScenes datasets. Particularly on the NuScenes dataset, at t = 3, the L2 (m) error decreases to 0.84 m, the collision rate (Col. Rate) drops to 0.22\%, and the CEGR (avg) reaches 4.89. The ECSA module establishes unified observation perspectives across tasks through equivariant operations, ensuring consistency in the prior observational data $O$ between motion prediction governed by $P(\tau \mid O, s_{t})$ and planning prediction governed by $P(\tau, s_{t} \mid O)$. With the introduction of the \textbf{UTTD} module, the overall performance of FUMP has been further significantly improved. On the NuScenes dataset and Bench2Drive, the avg of CEGR (L2) reached 14.68 and 5.15, respectively, demonstrating the model’s strong capability in motion learning and trajectory prediction. Compared to conventional approaches, UTTD achieves superior performance by reformulating the planning task as a multi-stage process that inherently incorporates motion prediction, thereby unifying the motion-planning pipeline. 


\textbf{On the NuScens-LT and Bench2Drive-LT Datasets (Open-Loop)}. Table~\ref{tab:ablation_fs} presents the ablation study results of the FUMP model on the NuScenes-LT and Bench2Drive-LT datasets.The \textbf{ESCA} employs equivariant operations to unify the data perspectives of motion and planning tasks. Building upon UMP, it achieves performance gains of 0.60m in average L2 error, 0.09\% in collision rate (on NuScenes-LT), and 2.03 in CEGR (avg). The \textbf{UTTD} unifies motion and planning while preserving task distinctions, delivering superior performance (avg) 0.87m (L2), 0.0\% (Col. Rate), 5.65 (CEGR) on Bench2Drive-LT. And the maintained optimal metrics of FUMP demonstrate that ECSA's data perspective transformation capability coupled with UTTD's task coordination synergy significantly enhance long-tail learning performance for both motion prediction and planning task.


\textbf{On the Bench2Drive (Close-Loop) Dataset}. As shown in Table~\ref{tab:ablation_close}, our systematic evaluation examines FUMP’s performance through four key metrics: Success Rate (SR), Drive Score (DS), and CEGR(DS)/CEGR(SR) that quantify motion data learning capabilities, route completion and driving quality. The unified approach, \textbf{UMP}, shows limited closed-loop effectiveness (SR: 38.87\%, DS: 13.63, CEGR (SR): -3.50, CEGR (DS):  -3.89), revealing fundamental limitations in handling  data distributional, and data perspective differences between motion and planning tasks. The ESCA module demonstrates measurable improvements (SR: 42.88\%, DS: 15.00, CEGR (SR): 0.53, CEGR (DS): 0.00), confirming its effectiveness in mitigating viewpoint variations for better generalization. Nobaly, The combined modules deliver state-of-the-art performance (SR: 45.67\%, DS: 16.36, CEGR (SR): 3.34, CEGR (DS): 3.86) and demonstrates synergistic benefits deriving from ESCA’s and UTTD’s structured task unification for reliable real-world driving.

\textbf{On the NAVSIM and NAVSIM-LT Dataset (Close-Loop).} As shown in Tabble~\ref{tab_ablation_nav} and~\ref{tab_ablation_nav_fs}, we systematically evaluate the contribution of each FUMP module. The UMP module provides limited performance gains, improving the PDMS score by only 1.5 and 0.5 on NAVSIM and NAVSIM-LT, respectively. This suggests that naively unifying motion and planning data during training fails to fully exploit the potential benefits of motion data for planning. In contrast, the ECSA module—which bridges the gap between motion and planning perspectives—delivers more substantial improvements. Most notably, the UTTD module achieves the highest gains, boosting PDMS scores by 1.8 and 2.8 on NAVSIM and NAVSIM-LT. This stems from UTTD’s explicit modeling of task discrepancies between motion and planning. By decomposing and aligning their shared representations, UTTD enables the planning task to fully leverage motion data while mitigating interference from task-specific variations.


\textbf{On the ADV-NuScenes Dataset (Open-Loop).} The experimental results in Table~\ref{tab_ablation_adv} reveal similar key insights about the module contributions in FUMP. While the standalone UMP module leads to a slight 2.8\% increase in average collision rate (from 1.026\% to 1.055\%), it demonstrates significant value in long-term safety by reducing the 3-second horizon collision probability by 14.0\%, highlighting its effectiveness in far-horizon collision anticipation. The ECSA module further enhances overall system performance, achieving a 6.2\% reduction in average collision rate along with a positive CEGR improvement of +2.86, confirming that its cross-view alignment mechanism successfully mitigates perception conflicts that typically cause collisions. The complete FUMP framework integrating all three modules establishes SOTA performance, delivering a 29.2\% lower average collision rate. Particularly noteworthy is the system's exceptional stability in obstacle avoidance at critical 3-second horizons, confirming the synergistic benefits of the integrated architectural design.

\subsubsection{The effect of hyper-parameters in FUMP}
As demonstrated in Tab.~\ref{tab_length} and Tab.~\ref{tab_gamma}, we systematically investigate the impact of hyperparameters $Length$ and $\gamma$ on model performance. The results reveal that increasing length from 100 to 500 leads to consistent performance gains across all datasets, while further extension to 700 shows diminishing returns. This observation aligns with our hypothesis that expanding the queue size provides richer references for planning tasks, thereby improving performance until reaching saturation. Interestingly, the model exhibits distinct behaviors with different $\gamma$ values. This because, when $\gamma$ is small (0.2), the queue update threshold depends more on current batch losses, whereas larger $\gamma$ values (0.4) depends more on historical loss. Our experiments show a slight performance degradation when increasing $\gamma$ from 0.2 to 0.4, indicating that smaller $\gamma$ values are preferable as training progresses and losses decrease, since they better preserve hard samples for queue updates.




\subsection{Robustness Study}
To rigorously evaluate the model's robustness in realistic noisy environments, we conduct comprehensive experiments across three major autonomous driving benchmarks, NuScenes, Bench2Drive, and NAVSIM (their results are showed in Table~\ref{noise_nus}, ~\ref{noise_ben} and \ref{noise_nav}). Our study incorporates five distinct noise types commonly encountered in real-world driving scenarios, motion blur (simulating camera shake), Gaussian noise (sensor imperfections), shear distortion (perspective variations), scale fluctuations (object size inconsistencies), and uniform noise (general environmental interference). The results reveal several critical findings. First, even minor noise injection leads to measurable performance degradation across all benchmarks, particularly in collision rate on NuScenes and EP score on NAVSIM. This sensitivity analysis confirms the necessity of explicit robustness considerations in autonomous driving systems. Notably, FUMP demonstrates superior noise resilience compared to baseline methods (SparseDrive, VAD, and TransFuser), maintaining robust performance across multiple challenging noise conditions. For Gaussian and Scale distortions in NAVSIM, FUMP has 21.1\% and 64.0\% lower performance degradation respectively compared to the best baseline; For Motion Blur in NuScenes, it achieves a 39.1\% reduction in collision rate relative to SparseDrive. These results conclusively demonstrate FUMP's exceptional robustness against sensor noise and its strong scalability across diverse noisy driving scenarios. 

\section{Visualization} \label{sec:visualization}
\begin{figure*}[]
    \centering
    \includegraphics[width=\linewidth]{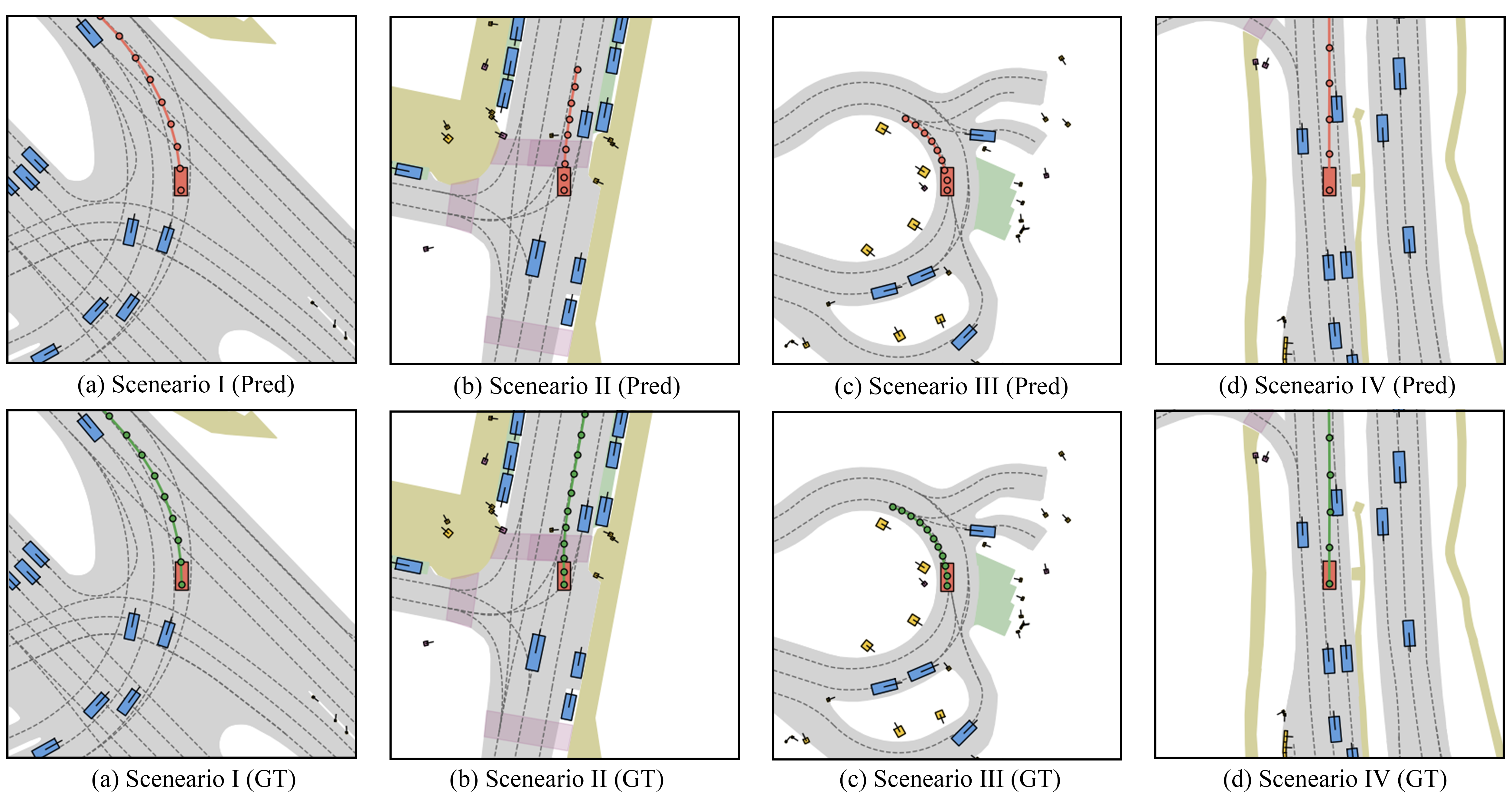}
    \caption{\textbf{The visualization results of FUMP with GT on NAVSIM-LT Dataset}. Four scenarios encompass: turning (a), lane change (b), loop (c), and overtaking (d).}
    \label{fig:vis_nav}
\end{figure*}

We visualize FUMP’s performance on NAVSIM-LT, focusing on long-tail driving scenarios. For NAVSIM-LT, we select four scenarios that includes turning, lane change, loop, and overtaking. As shown in Fig.~\ref{fig:vis_nav}, FUMP achieves robust performance in all data sets of a long-tails, particularly in overtaking and lane change scenarios. This highlights its capability to effectively transfer knowledge from other vehicles’ data to enhance its own driving skills.

\section{Conclusion} \label{sec:conclusion}
In this work, we propose FUMP, an end-to-end autonomous driving framework that leverages motion data to enhance the model’s learning capacity and improve long-tail performance. Specifically, FUMP employs a graph-based equivariant structure to unify the observational perspectives of both ego and other vehicles, ensuring cross-vehicle viewpoint consistency. We further introduce a novel decoder architecture to jointly optimize motion prediction and planning tasks. For evaluation, we construct a long-tail benchmark by subsampling NuScenes, Bench2Drive and NAVSIM validation sets. Extensive experiments demonstrate that FUMP significantly outperforms state-of-the-art end-to-end methods in trajectory prediction, especially at long-tail settings. Our work aims to advance the community’s understanding of motion data utilization and its benefits for planning tasks.

\textbf{Limitation and future work.} Although FUMP effectively unifies motion planning and trajectory optimization tasks, its vehicle state estimator demonstrates limited reliability in handling occluded or sensor-missing scenarios, particularly for surrounding vehicles with incomplete modality inputs. This limitation motivates our future research focus on robust multi-agent state estimation under perceptual uncertainty.  

\section*{ACKNOWLEDGMENTS} 
This work was supported in part by the National Key R\&D Program of China (2018AAA0100302), supported by the STI 2030-Major Projects under Grant 2021ZD0201404.

\ifCLASSOPTIONcaptionsoff
  \newpage
\fi

\bibliographystyle{IEEEtran}
\bibliography{egbib}

\begin{IEEEbiography}
[{\includegraphics[width=1in,height=1.25in,clip,keepaspectratio]{{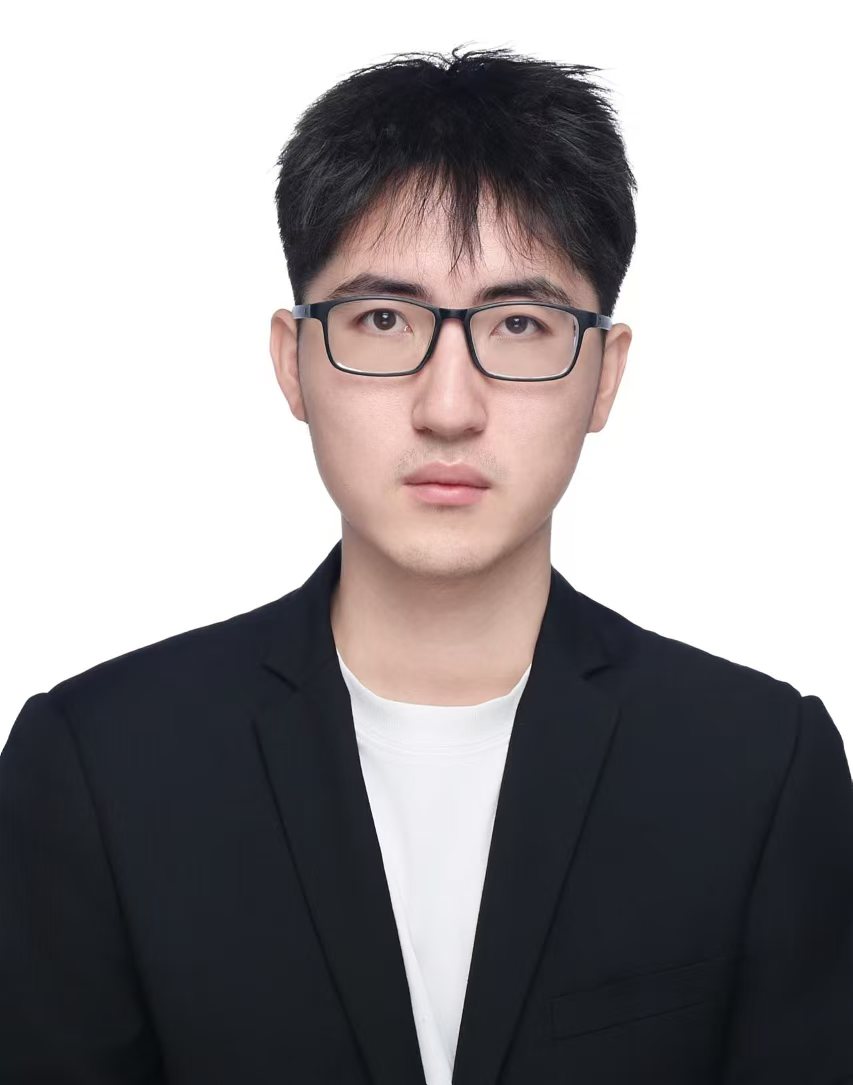}}}] {Lin Liu} was born in Jinzhou, Liaoning Province, China, in 2001. He is now a college student majoring in Computer Science and Technology at China University of Geosciences(Beijing).
Since Dec. 2022, he has been recommended for a master's degree in Computer Science and Technology at Beijing Jiaotong University. His research interests are in computer vision.
\end{IEEEbiography}

\begin{IEEEbiography}[{\includegraphics[width=1in,height=1.25in,clip,keepaspectratio]{{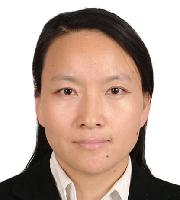}}}]{Caiyan Jia}, born on March 2, 1976, is a lecturer and a postdoctoral fellow of the Chinese Computer Society. she graduated from Ningxia University in 1998 with a bachelor's degree in mathematics, Xiangtan University in 2001 with a master's degree in computational mathematics, specializing in intelligent information processing, and the Institute of Computing Technology of the Chinese Academy of Sciences in 2004 with a doctorate degree in engineering, specializing in data mining. she has received her D. degree in 2004. She is now a
professor in School of Computer Science and
Technology, Beijing Jiaotong University, Beijing,
China.
\end{IEEEbiography}

\begin{IEEEbiography}[{\includegraphics[width=1in,height=1.25in,clip,keepaspectratio]{{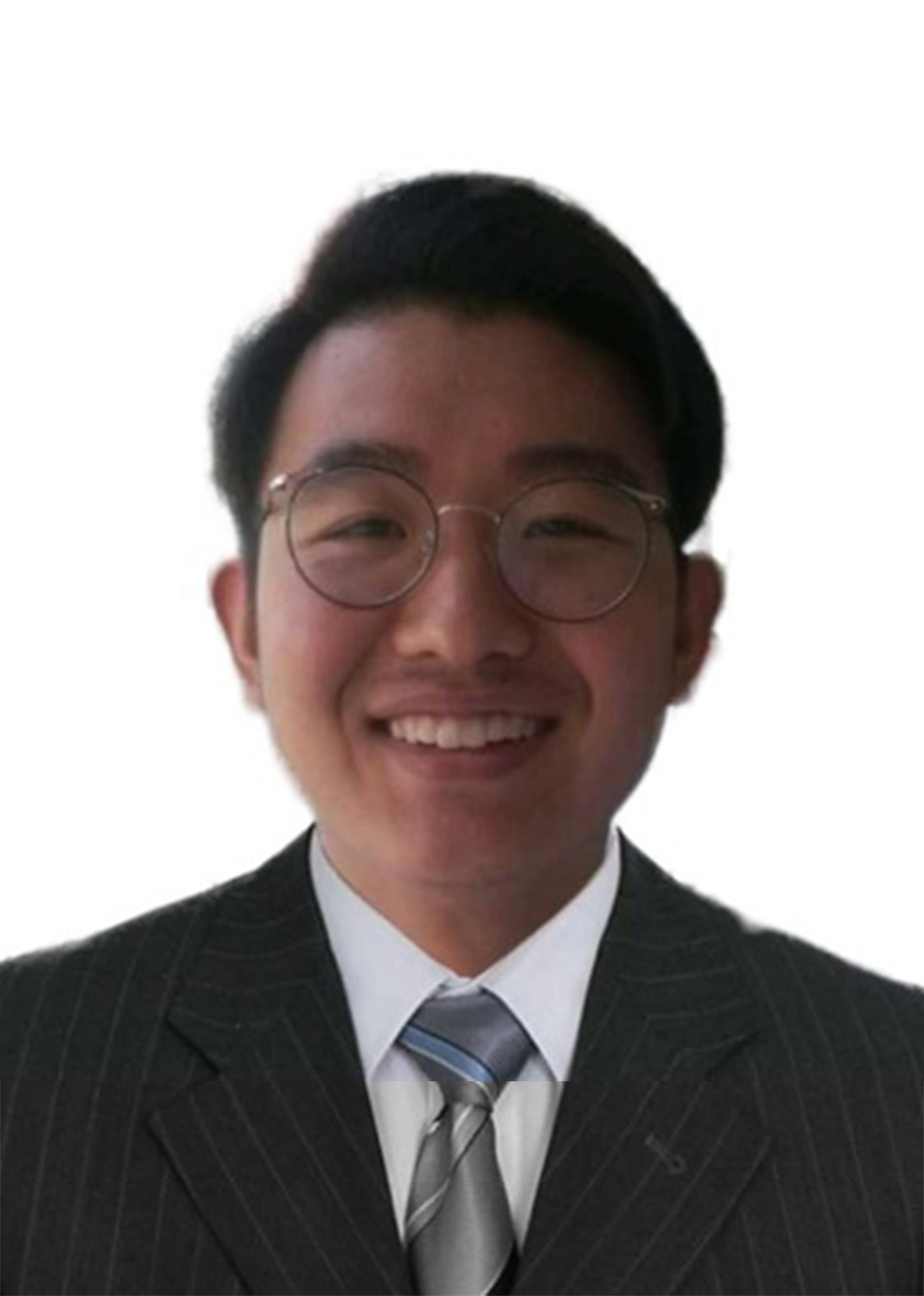}}}]{Ziying Song} was born in Xingtai, Hebei Province, China in 1997. He received the B.S. degree from Hebei Normal University of Science and Technology (China) in 2019. He received a master's degree major in Hebei 
University of Science and Technology (China) in 2022. He is now a PhD student majoring in Computer Science and Technology at Beijing Jiaotong University (China), with a research focus on Computer Vision. 
\end{IEEEbiography}

\begin{IEEEbiography}[{\includegraphics[width=1in,height=1.25in,clip,keepaspectratio]{{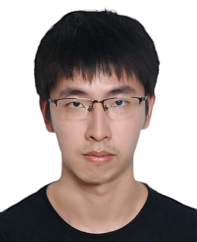}}}]{Hongyu Pan}  received the B.E. degree from Beijing Institute of Technology (BIT) in 2016 and the M.S. degree in computer science from the Institute of Computing Technology (ICT), University of Chinese Academy of Sciences (UCAS), in 2019. He is currently an employee at Horizon Robotics. His research interests include computer vision, pattern recognition, and image processing. He specifically focuses on 3D detection/segmentation/motion and depth estimation.

\end{IEEEbiography}

\begin{IEEEbiography}[{\includegraphics[width=1in,height=1.25in,clip,keepaspectratio]{{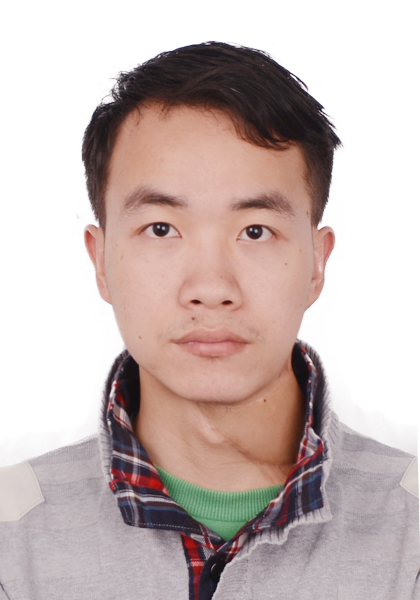}}}]{Bencheng Liao} received the B.E. degree from School of Electronic Information and Communications, Huazhong University of Science and Technology, Wuhan, China, in 2020. He is currently a PhD candidate at the Institute of Artificial Intelligence and School of Electronic Information and Communications, Huazhong University of Science and Technology. His research interests include object detection, 3D vision, and autonomous driving.
\end{IEEEbiography}

\begin{IEEEbiography}[{\includegraphics[width=1in,height=1.25in,clip,keepaspectratio]{{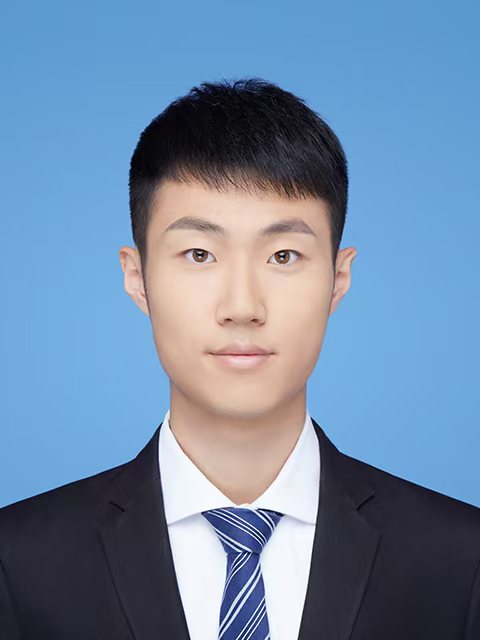}}}]{Wenchao Sun} 
is a Ph.D. candidate working on computer vision and deep learning at Beijing Key Lab of Traffic Data Analysis and Mining, Beijing Jiaotong University as of 2020. Before that he received the B.S. degree at Beijing Jiaotong University in 2020. He is about to work as a researcher at Baidu Inc. His research interest includes visual object tracking, BEV detection/tracking and E2E-AD.
\end{IEEEbiography}

\begin{IEEEbiography}[{\includegraphics[width=1in,height=1.25in,clip,keepaspectratio]{{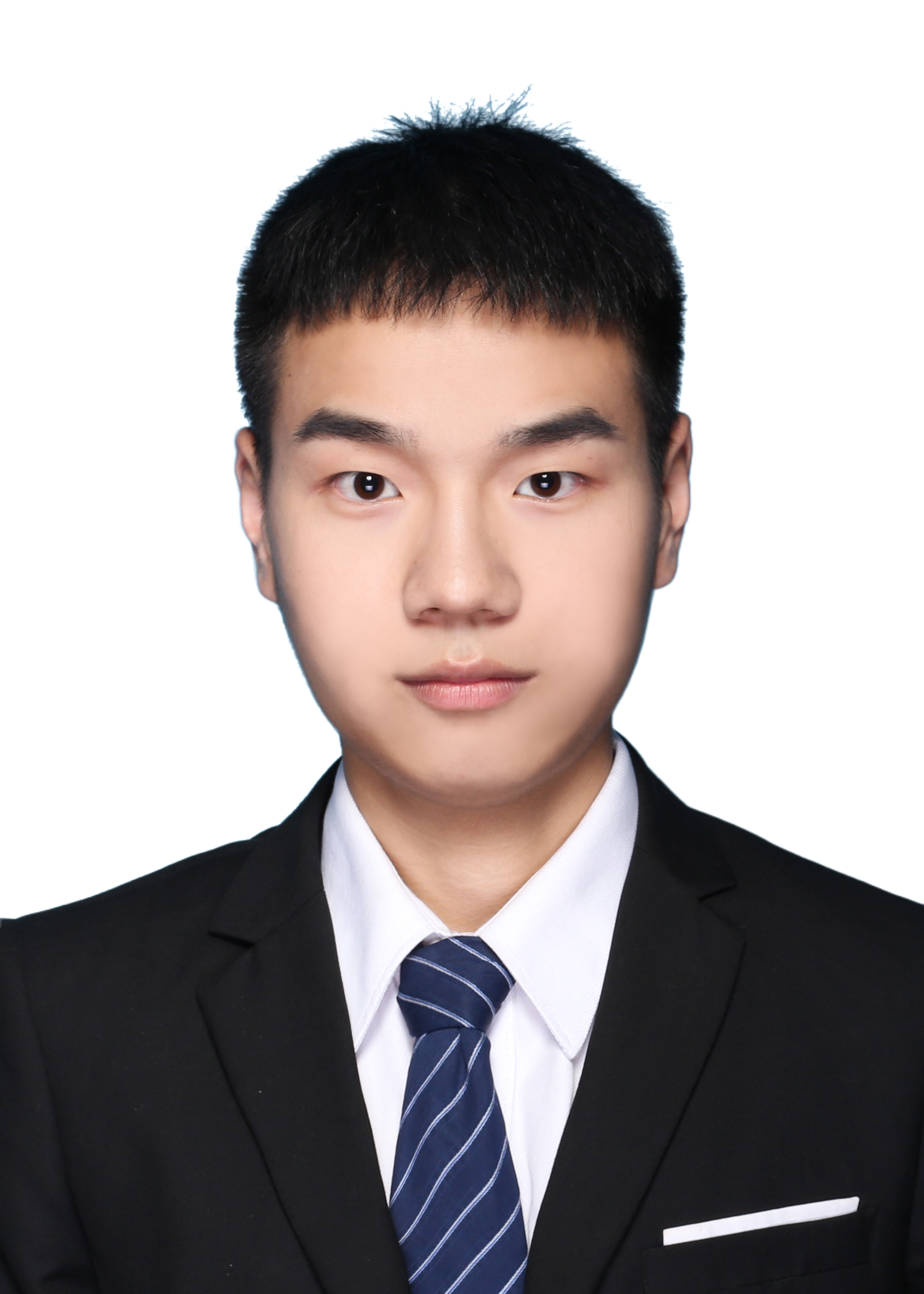}}}]{Yongchang Zhang} received the B.E. degree from the University of Electronic Science and Technology of China (UESTC) in 2020 and the M.S. degree in control theory and control engineering from the Institute of Automation (CASIA), University of Chinese Academy of Sciences (UCAS) in 2023. He is currently a Research Engineer at Horizon Robotics. His research interests include computer vision, autonomous driving perception, and visual-language models. \end{IEEEbiography}

\begin{IEEEbiography}[{\includegraphics[width=1in,height=1.25in,clip,keepaspectratio]{{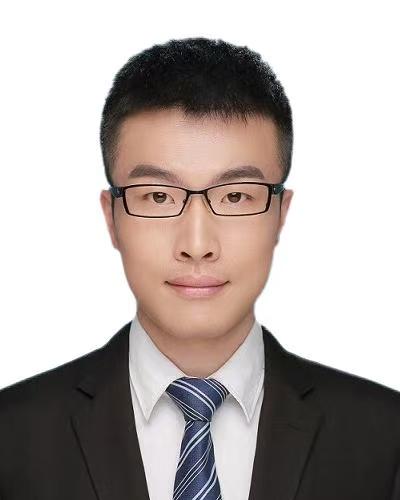}}}]
{Lei Yang (Member, IEEE)} received his M.S. degree from the Robotics Institute at Beihang University, in 2018. and the Ph.D. degree from the School of Vehicle and Mobility, Tsinghua University, in 2024. From 2018 to 2020, he joined the Autonomous Driving R\&D Department of JD.COM as an algorithm researcher. Currently, he is a research fellow with the School of Mechanical and Aerospace Engineering, Nanyang Technological University, Singapore. His current research interests include autonomous driving, 3D scene understanding and world model.
\end{IEEEbiography}

\begin{IEEEbiography}[{\includegraphics[width=1in,height=1.25in,clip,keepaspectratio]{{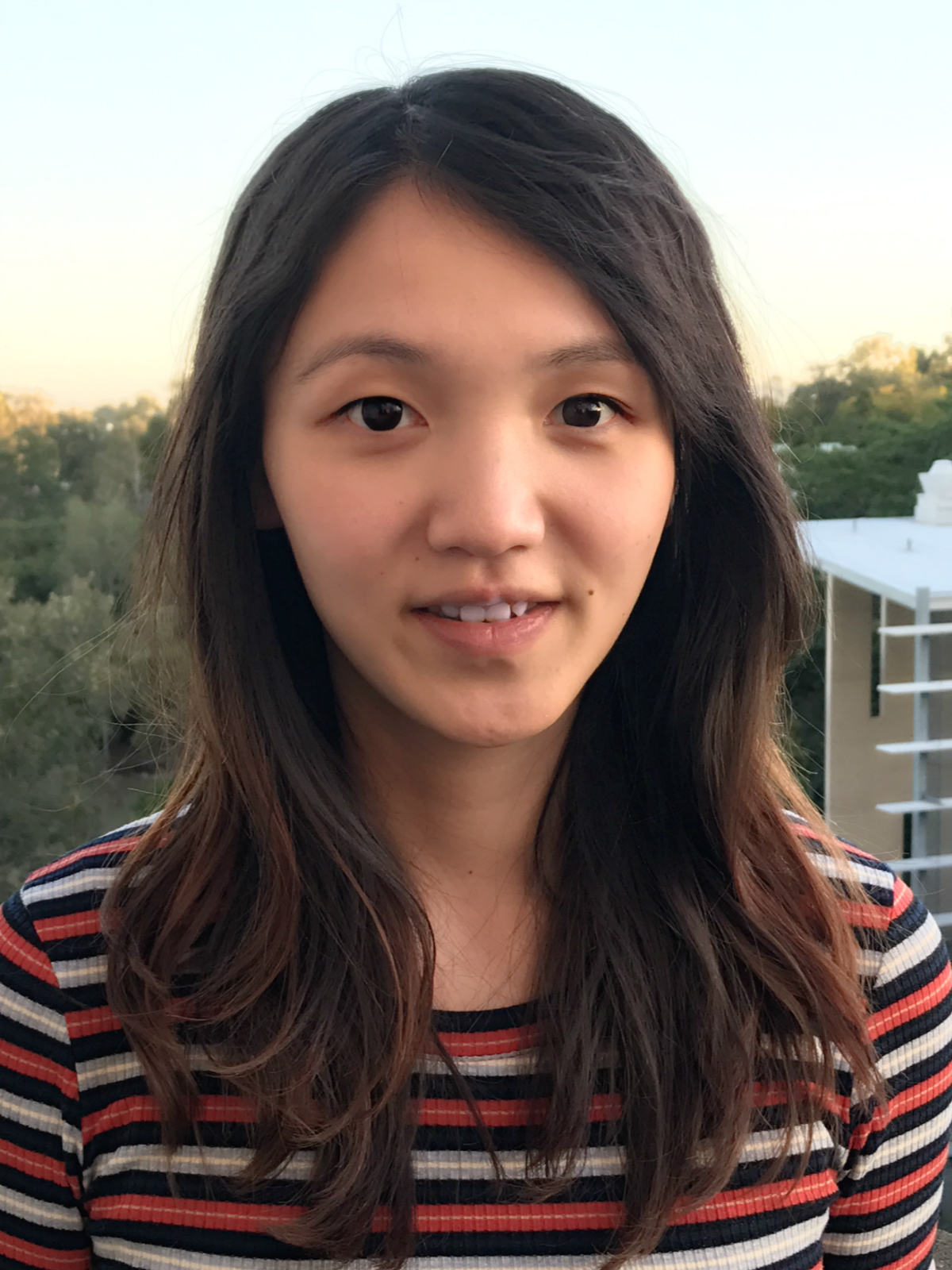}}}]{Yadan Luo} (Member, IEEE) received the BS degree in computer science from the University of Electronic Science and Technology of China, and the PhD degree from the University of Queensland. Her research interests include machine learning, computer vision, and multimedia data analysis. She is now a lecturer with the University of Queensland.
\end{IEEEbiography}

\end{document}